\pdfoutput=1
\documentclass[11pt]{article}

\usepackage[preprint]{acl}
\usepackage{times}
\usepackage{latexsym}
\usepackage[T1]{fontenc}
\usepackage[utf8]{inputenc}

\usepackage{microtype}

\usepackage{inconsolata}

\usepackage{graphicx}
\usepackage{amsmath}
\usepackage{booktabs}
\usepackage{array}

\usepackage{wrapfig}
\usepackage{inconsolata}
\usepackage{xspace}
\usepackage{graphicx}
\usepackage{booktabs} 
\usepackage{marvosym}  % Package for symbols
\usepackage{graphicx}
\usepackage{caption}
\usepackage{stfloats} % add this to your preamble
\usepackage{cuted}
\usepackage{multirow}
\usepackage{tabularx}
\usepackage{subcaption}
\usepackage{flushend}

\usepackage{algorithm}
\usepackage{algpseudocode}

\newcommand\CONDITION[2]%
   {\begin{tabular}[t]{@{}l@{}}#1#2\end{tabular}}
\algdef{SE}[WHILE]{While}{EndWhile}[1]%
   {\algorithmicwhile\ \CONDITION{#1}{\ \algorithmicdo}}%
   {\algorithmicend\ \algorithmicwhile}
\algdef{SE}[FOR]{For}{EndFor}[1]%
   {\algorithmicfor\ \CONDITION{#1}{\ \algorithmicdo}}%
   {\algorithmicend\ \algorithmicfor}
\algdef{S}[FOR]{ForAll}[1]%
   {\algorithmicforall\ \CONDITION{#1}{\ \algorithmicdo}}
\algdef{SE}[REPEAT]{Repeat}{Until}%
   {\algorithmicrepeat}[1]%
   {\algorithmicuntil\ \CONDITION{#1}{}}
\algdef{SE}[IF]{If}{EndIf}[1]%
   {\algorithmicif\ \CONDITION{#1}{\ \algorithmicthen}}%
   {\algorithmicend\ \algorithmicif}
\algdef{C}[IF]{IF}{ElsIf}[1]%
   {\algorithmicelse\ \algorithmicif\ \CONDITION{#1}{\ \algorithmicthen}}
\makeatletter
\ifthenelse{\equal{\ALG@noend}{t}}%
   {\algtext*{EndWhile}%
    \algtext*{EndFor}%
    \algtext*{EndLoop}%
    \algtext*{EndIf}%
   }{}%
\makeatother

\usepackage{tocloft}
\usepackage{etoc}

\usepackage{adjustbox}
\usepackage{enumitem}

\usepackage[table]{xcolor}
\usepackage{lipsum}
\usepackage{titletoc}
\usepackage{tcolorbox}
\usepackage{arydshln}
\usepackage{comment}

\tcbuselibrary{breakable}

\newcommand{\ourwork}{\textsc{HiRAS}\xspace}

    \newcommand{\darkred}[1]{\textcolor{red!75!black}{#1}}

\newif\ifreview
\reviewfalse     % Final 模式

\newcommand{\hh}[1]{%
\ifreview
{\textcolor{orange!90!black}{#1}}
\else
#1
\fi
}

\newcommand{\tcbtab}{\hspace*{4ex}}
\newcommand{\hyphentt}[1]{\texttt{\hyphenchar\font=\defaulthyphenchar #1}}

\algrenewcommand\algorithmicrequire{\textbf{Input:}}
\algrenewcommand\algorithmicensure{\textbf{Output:}}

\expandafter\def\expandafter\normalsize\expandafter{%
    \normalsize%
    \setlength\abovedisplayskip{-9pt}%
    \setlength\belowdisplayskip{4pt}%
    \setlength\abovedisplayshortskip{-8pt}%
    \setlength\belowdisplayshortskip{2pt}%
}
\title{\ourwork: A Hierarchical Multi-Agent Framework for \\ Paper-to-Code Generation and Execution}

\author{
 \textbf{Hanhua Hong\textsuperscript{1,2}},
 \textbf{Yizhi Li\textsuperscript{1}},
 \textbf{Jiaoyan Chen\textsuperscript{1}},
\\
 \textbf{Sophia Ananiadou\textsuperscript{1,3}},
 \textbf{Xiaoli Li\textsuperscript{4}},
 \textbf{Jung-jae Kim\textsuperscript{2}},
 \textbf{Chenghua Lin\textsuperscript{1}}
\\
 \textsuperscript{1}The University of Manchester, 
 \textsuperscript{2}Institute for Infocomm Research (I²R), A*STAR
 \\
 \textsuperscript{3}ELLIS Manchester, 
 \textsuperscript{4}Singapore University of Technology and Design
 \\
 \texttt{hanhua.hong@postgrad.manchester.ac.uk}
 \\
 \texttt{xiaoli\_li@sutd.edu.sg, kim\_jung\_jae@a-star.edu.sg}
  \\
 \texttt{\{yizhi.li-2,jiaoyan.chen,sophia.ananiadou,chenghua.lin\}@manchester.ac.uk}
}

\begin{document}

% TODO: Detailed Captions
    \maketitle
\begin{abstract}
Recent advances in large language models have highlighted their potential to automate computational research, particularly reproducing experimental results.
%, which is increasingly important given the limited proportion of released codebases relative to the growing number of conference publications. 
However, existing approaches still use fixed sequential agent pipelines with weak global coordination, which limits their robustness and overall performance. In this work, we propose Hierarchical Research Agent System (\ourwork), a hierarchical multi-agent framework for end-to-end experiment reproduction that employs supervisory manager agents to coordinate specialised agents across fine-grained stages. We also identify limitations in the reference-free evaluation of the Paper2Code benchmark and introduce Paper2Code-Extra (P2C-Ex), a refined protocol that incorporates repository-level information and better aligns with the original reference-based metric. We conduct extensive evaluation, validating the effectiveness and robustness of our proposed methods, and observing improvements, including >10\% relative performance gain beyond the previous state-of-the-art using open-source backbone models
%state-of-the-art results increasing over 10\% achieved with an open-source backbone model 
and significantly reduced hallucination in evaluation. 
%All code and data will be made publicly available. %[Github/HF Link]
Our work is available on GitHub: \url{https://github.com/KOU-199024/HiRAS}.
\end{abstract}

\section{Introduction} 

Recent advances in artificial intelligence (AI), particularly large language models (LLMs), have demonstrated remarkable capability in code generation and software development~\cite{9849664,yetiştiren2023evaluatingcodequalityaiassisted,openai2024gpt4ocard}. These developments have motivated a growing amount of research exploring the use of AI systems to assist scientific research, especially within Computer Science~\cite{chen2025ai4researchsurveyartificialintelligence,eger2025transforming}. AI Scientist~\cite{DBLP:journals/corr/abs-2408-06292} is one of the earliest attempts, which aims to automate the entire scientific workflow. Beyond such all-in-one pipelines, a substantial number of studies focus on individual tasks in the research lifecycle, including idea generation~\cite{wang-etal-2024-scimon,li-etal-2025-chain-ideas,garikaparthi-etal-2025-iris}, paper review~\cite{DBLP:journals/corr/abs-2402-10886,james-etal-2024-rigour,zhu2025deepreviewimprovingllmbasedpaper}, and experiment conduction~\cite{10.5555/3692070.3692738,schmidgall-etal-2025-agent}.

Among these tasks, reproduction of experimental pipelines and results described in published papers is of particular importance, since cumulative scientific progress fundamentally depends on the reproducibility of prior works~\cite{resnik2017reproducibility,10.5555/3546258.3546422,james2026rigourate}. However, the exponential growth in the number of research papers, combined with the fact that only around 20\% of papers  release complete and usable code repositories on average, has made exhaustive reproduction increasingly impractical for  researchers~\cite{doi:10.1142/S0218194022500358,magnusson-etal-2023-reproducibility,seo2025paper2codeautomatingcodegeneration}. 
This limitation naturally motivates the development of LLM-based agents to automate experiment reproduction. To evaluate how well existing systems can reproduce experimental pipelines and results directly from research papers, several benchmarks, such as PaperBench~\cite{starace2025paperbenchevaluatingaisability} and Paper2Code~\cite{seo2025paper2codeautomatingcodegeneration}, have been proposed. \hh{Generally, they use LLM-as-a-judge evaluation metrics due to the complexity of the task.} Results on these benchmarks indicate that, despite strong general-purpose coding ability, directly prompted LLMs achieve only limited success~\cite{xiang2025scireplicatebench,kim2025reproductionreplicationevaluatingresearch}. 
%This limitation naturally motivates the development of LLM-based agents to assist in automating the reproduction of experimental pipelines and results. To assess current capabilities, several benchmarks, \hh{such as PaperBench~\cite{starace2025paperbenchevaluatingaisability} and Paper2Code~\cite{seo2025paper2codeautomatingcodegeneration}}, have been proposed to evaluate whether AI systems can reproduce experimental pipelines and results directly from research papers. These studies show that, though possessing strong general-purpose coding ability, directly prompting LLMs achieves only limited success~\cite{xiang2025scireplicatebench,kim2025reproductionreplicationevaluatingresearch}.

To address this gap, recent works have adopted multi-agent frameworks that decompose the reproduction process into multiple stages. For example, PaperCoder~\cite{seo2025paper2codeautomatingcodegeneration} structures reproduction into planning, analysis, and coding phases, while AutoReproduce~\cite{zhao2025autoreproduceautomaticaiexperiment} combines search agents and coding agents to retrieve relevant studies and implement experimental code. 
%Despite these advances, existing systems largely arrange agents into a fixed sequential pipeline, implemented as fixed prompt invocations with limited global oversight. 
%As a result, errors introduced by one agent can easily propagate, leading to stalled workflows and incomplete repositories. 
Despite these advances, existing systems largely rely on fixed, sequential agent pipelines implemented via fixed prompt invocations with limited global supervision. Consequently, errors introduced by agents early in the pipeline  can easily propagate, resulting in stalled workflows and incomplete repositories. 
This lack of adaptive coordination fundamentally limits robustness, fault tolerance, and scalability in complex experimental environments~\cite{xie2025faraiscientistschanging,wei2025browsecompsimplechallengingbenchmark}.
%This lack of adaptive coordination inherently limits robustness, fault tolerance, and scalability in the complex environment of experiment
In contrast to fixed sequential pipelines, hierarchical agent architectures have been shown to be effective across domains such as embodied intelligence~\cite{lallement2014hatphtnplannerrobotics}, reinforcement learning~\cite{wang2020romamultiagentreinforcementlearning}, and energy systems~\cite{dragomir2025decentralized}. By introducing explicit manager roles, hierarchical organisation enables effective communication across the system, thereby improving cooperation and scalability in complex multi-agent systems~\cite{feng2024hierarchicalconsensusbasedmultiagentreinforcement,mooretaxonomy}. 
%\hh{Inspired by these successes, we adopt a hierarchical architecture with manager agents for paper reproduction to strengthen quality supervision and mitigate error propagation throughout the workflow.}
%\hh{Therefore, we incorporate a hierarchical system with manager agents in paper reproduction to enhance quality supervision and mitigate error propagation across the process.}
%\red{[CL: try to better highlight what is novel and distinctive about our work; only mentioning that hierarchical orchestration has not been employed in the paper domain is a bit thin.]}

In this work, we propose \textbf{Hi}erarchical \textbf{R}esearch \textbf{A}gent \textbf{S}ystem (\textbf{\ourwork}), a novel multi-agent framework that introduces dedicated manager agents to supervise and control the global experiment reproduction process. Unlike prior hierarchical multi-agent systems, where manager agents  passively deliver messages~\cite{DBLP:journals/corr/abs-1901-08492, wu2024autogen}, the managers in \ourwork function as proactive supervisors, which actively inspect task progress and dynamically invoke specialised subordinate agents to produce artefacts and correct errors. For example, a global manager may invoke a coding agent to implement experimental code and re-invoke it when the subsequent execution agent reports errors in the program. 
In addition, the reproduction workflow is decomposed into finer-grained phases for these specialised agents, which are equipped with appropriate tools to interact with a shared workspace. Together, these design choices enable effective collaboration and information exchange across the system. To the best of our knowledge, \ourwork is the first framework to incorporate global, supervisory manager agents in a hierarchical multi-agent architecture for automated  experiment reproduction.

%Nevertheless, such hierarchical orchestration has not yet been systematically explored for end-to-end research paper reproduction.
%Unlike prior hierarchical multi-agent systems, where manager agents often passively deliver messages~\cite{DBLP:journals/corr/abs-1901-08492, wu2024autogen}, the managers in \ourwork function as \textcolor{orange}{proactive} supervisors. 

%To validate the effectiveness of \ourwork,
For evaluation, we conduct comprehensive experiments on the PaperBench and Paper2Code benchmarks using two popular backbone models, \hyphentt{Qwen3-Coder-480B}~\cite{yang2025qwen3technicalreport} and \hyphentt{DeepSeek-v3.1-Terminus}~\cite{deepseekai2025deepseekv3technicalreport}. 
Our approach consistently outperforms baselines across all settings when using the same backbone model. Specifically, on PaperBench, \ourwork even outperforms the current state-of-the-art methods which use proprietary LLMs by a substantial margin.
%and even achieves \textit{state-of-the-art} (SOTA) results} on PaperBench with an open-source backbone model, demonstrating both effectiveness and generalisability. 
On Paper2Code, our analysis further identifies fundamental limitations of the reference-free \hh{prompt-based} evaluation protocol introduced in the original work~\cite{seo2025paper2codeautomatingcodegeneration}, particularly its susceptibility to evaluator hallucination. To address this issue, we introduce \textbf{Paper2Code-Extra (P2C-Ex)}, a refined reference-free evaluation paradigm that incorporates explicit repository-level information. Empirical results show that P2C-Ex substantially improves alignment with reference-based evaluation, offering a more reliable framework for assessing paper-to-code consistency in the absence of gold repositories.

To summarise, the contributions of our work are three-fold:
\setlist{nolistsep}
\begin{itemize}
    \item We propose \textbf{\ourwork}, a novel hierarchical multi-agent framework for automated experiment reproduction from research papers, which integrates specialised agents with supervisory manager agents for improved coordination.
    %a novel hierarchical multi-agent framework for automated research paper reproduction, which integrates specialised agents with supervisory manager agents for improved coordination.
    \item We evaluate our framework on two experiment reproduction benchmarks using three LLM backbones, demonstrating consistent and substantial improvements over prior approaches and achieving state-of-the-art performance. 
    %We conduct experiments on two paper reproduction benchmarks using two backbone LLMs, demonstrating consistent and substantial improvements over prior approaches, achieving SOTA performance with open-source models.
    \item We introduce \textbf{P2C-Ex}, a reference-free evaluation protocol for the Paper2Code benchmark that incorporates repository-level information and yields stronger alignment with reference-based metrics than the original protocol of Paper2Code.
\end{itemize}

\begin{figure*}
    \centering
    \includegraphics[width=0.85\textwidth]{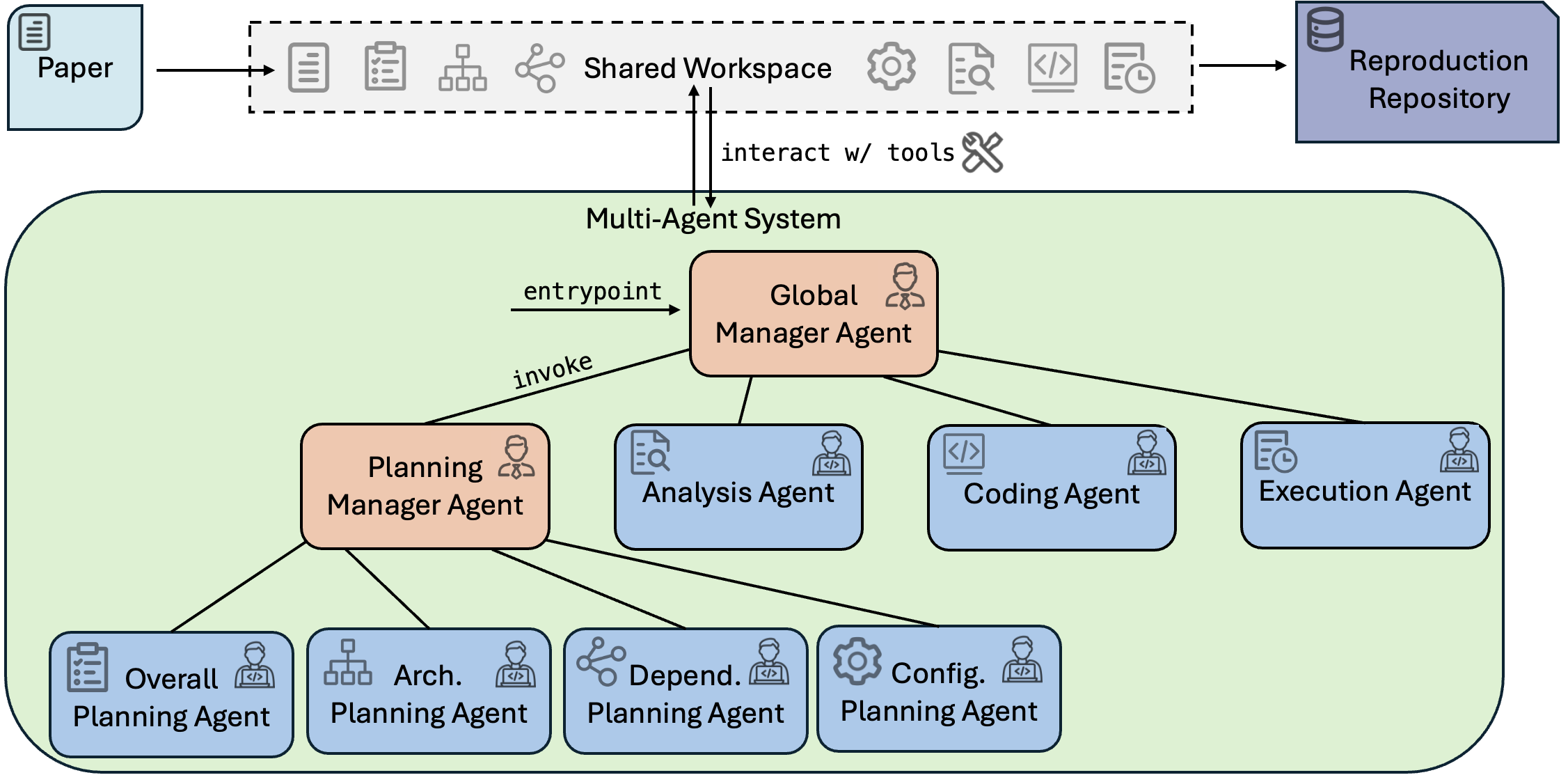}
    \caption{Overview of the \ourwork framework. The reproduction workflow is decomposed into fine-grained phases, each handled by a specialised agent (blue) equipped with appropriate tools to operate within a shared workspace throughout the process. To enhance coordination, \ourwork introduces hierarchical manager agents (orange) that inspect the workspace to supervise progress and dynamically invoke subordinate agents to perform tasks and correct errors according to the feedback results.}
    \label{fig:main}
    \vspace{-1.5em}
\end{figure*}
%TODO: Add more illustration about the figure

\section{Related Work} 

%Our work aims to advance AI systems for computer science research through the development of a multi-agent system with hierarchical orchestration. In this section, we will review prior related works and clearly delineate how our approach differs from existing methods, thereby establishing its novel contributions.

\subsection{AI for Scientific Research}
The rapid advancement of LLMs has resulted in their increasing application in scientific research, particularly within the computer science domain~\cite{gottweis2025aicoscientist,xie2025faraiscientistschanging,si2025can,wu2026largescaleterminalagentictrajectory}. Some studies explore the use of LLMs to perform the entire scientific research lifecycle, including the AI Scientist~\cite{DBLP:journals/corr/abs-2408-06292,yamada2025aiscientistv2workshoplevelautomated}, DOLPHIN~\cite{yuan-etal-2025-dolphin}, and Zochi~\cite{zochi2025}, which produced papers accepted at major conferences.

In addition to such holistic frameworks, other work has focused on individual components of the research process. For example, SciMON~\cite{wang-etal-2024-scimon} and Chain-of-Ideas~\cite{li-etal-2025-chain-ideas} improve the generation of AI-driven ideas, whilst other frameworks have been developed to identify and assess the scientific rigour described in research papers \citep{james-etal-2024-rigour}.
Another key direction is experiment reproduction, in which models reproduce experimental results from published studies. As benchmarks, SciReplicateBench~\cite{xiang2025scireplicatebench} evaluates reproducibility by requiring agents to complete missing code segments, and PaperBench~\cite{starace2025paperbenchevaluatingaisability} provides detailed, paper-specific rubrics for assessing end-to-end reproduction fidelity. To support reproduction, PaperCoder~\cite{seo2025paper2codeautomatingcodegeneration} reconstructs experimental codebases from published papers using multi-stage task decomposition, and AutoReproduce~\cite{zhao2025autoreproduceautomaticaiexperiment} further enhances reproduction by automatically retrieving relevant literature and associated code repositories.

However, prior work on experiment reproduction has largely relied on fixed, sequential workflows, in which errors introduced by early-stage agents propagate unchecked. In contrast, our approach introduces a hierarchical management architecture that provides explicit supervisory control: manager agents actively inspect intermediate artefacts, diagnose failures, and re-invoke stage-specific agents to correct errors throughout the reproduction process.

%However, prior work on paper reproduction has largely been constrained to fixed sequential workflows. In contrast, our approach employs a hierarchical management architecture where manager agents actively supervise experimental progress and coordinate stage-specific agents, providing instructions throughout the reproduction process.

\subsection{Multi-Agent Systems}
Multi-agent systems have gained increasing attention due to the widespread adoption of AI agents across diverse applications~\cite{guo2024large,hong2025onesizefitsallinversionlearninghighly,li2024survey}. A key distinction among these systems is their coordination strategies, which determine how agents communicate and collaborate~\cite{sun2025llm, DBLP:journals/corr/abs-2502-14743}. While some adopt flat, decentralised schemes~\cite{10.5555/3692070.3692537,DBLP:journals/corr/abs-2502-14743}, hierarchical organisation has proven more effective for managing complexity and enhancing scalability~\cite{mooretaxonomy}. Examples include FMH~\cite{DBLP:journals/corr/abs-1901-08492}, where a manager delegates sub-goals to workers; \citet{liang-etal-2024-encouraging}, which uses a judge to summarise agent debates; and AutoGen~\cite{wu2024autogen}, where managers mediate and disseminate messages among agents.

Unlike prior works, where manager agents primarily act as passive communication intermediaries, our framework assigns them an active supervisory role. Managers oversee progress, inspect artefacts, and control the workflow by invoking phase-specific specialised agents. This enables effective global coordination throughout end-to-end experiment reproduction.

\section{Methodology}
\subsection{Problem Definition}
%In this work, we focus on an end-to-end paper reproduction task.
In this work, we use research agents for experiment reproduction.
A \textbf{research agent} is defined as a tuple $\mathcal{A}=(\mathrm{LLM},\mathrm{Mem},T,E,K)$, comprising a backbone model $\mathrm{LLM}$, an individual memory context $\mathrm{Mem}$ (including prompts), and access to a suite of tools $T$ that enable interaction with a workspace environment $E$. The agents follow the ReAct paradigm~\cite{yao2023react}, performing up to $K$ reasoning-action iterations, while optionally calling $\texttt{end\_task}$ to submit outputs early.

We use a system composed of multiple research agents defined above to deal with this task, denoted as $\mathcal{S}=\{\mathcal{A}_1,\mathcal{A}_2,\dots\}$.
%Due to the inherent complexity of this task, a multi-agent system $\mathcal{S}=\{\mathcal{A}_1,\mathcal{A}_2,\dots\}$ is often instantiated to complete the full reproduction. 
%In this way, 
Given a research paper $P$, the objective of the multi-agent system $\mathcal{S}$ is to generate the full reproduction output, expressed as $R = \mathcal{S}(P)$.

\subsection{Method Overview}
%To tackle these challenges, we propose \textbf{\ourwork}, a novel hierarchical multi-agent framework for automated paper reproduction. 
The workflow of \ourwork is decomposed into fine-grained phases, executed by specialised agents, while dedicated manager agents supervise progress and dynamically invoke their subordinate specialised agents. An overview of \ourwork is shown in Figure \ref{fig:main}. 

\subsection{Specialised Agents} 
%Following  PaperCoder~\cite{seo2025paper2codeautomatingcodegeneration}, we divide the reproduction workflow into three primary phases: \textit{planning}, \textit{analysis}, and \textit{coding}, 
In decomposing the reproduction workflow, we keep the three primary phases: \textit{planning}, \textit{analysis}, and \textit{coding}, following PaperCoder~\cite{seo2025paper2codeautomatingcodegeneration},
and introduce an additional \textit{execution} phase to validate code executability. The planning phase is further decomposed into sub-phases: overall planning, architecture design, dependency modelling, and configuration generation. Unlike PaperCoder, which employs only one agent for planning, we assign each sub-phase to a specialised agent: $\mathcal{A}_\mathrm{overall}, \mathcal{A}_\mathrm{arch}, \mathcal{A}_\mathrm{dep}, \mathcal{A}_\mathrm{config}$, to sequentially generate the plans. Then, we instantiate a specialised agent $\mathcal{A}_t$ for each remaining phase, yielding $\mathcal{A}_\mathrm{analysis}, \mathcal{A}_\mathrm{code}, \mathcal{A}_\mathrm{exec}$. Every agent is initialised with its initial memory context $\mathrm{Mem}_{t}$. 

All agents operate within a shared workspace $E$, initialised with only the paper $E = \{P\}$. To interact with this workspace, agents are equipped with a common file system toolset:

\begin{equation}
%\small
T_\mathrm{file} = \{\texttt{list\_dir}, \texttt{read\_file}, \texttt{write\_file}\},
\end{equation}
enabling them to store outputs and access both the paper and artefacts generated by other agents. The execution agent has an extended toolset: $
%T_{\mathrm{exec}}=
T_\mathrm{file}\cup\{\texttt{bash}\}$, enabling direct interaction with the system console for program execution.

Specifically, during the planning phase, agents sequentially generate a set of structured plans: an overall plan that summarises the experiments described in the paper; an architecture design that specifies the components required for implementation and their structure; a dependency plan that models the function-level relationships among these components; and a configuration plan that details the experimental parameters. Each plan is produced by a dedicated sub-phase agent, collectively forming the plan set: $
\mathrm{Pln}=\{\mathrm{Pln}_{\mathrm{overall}},\mathrm{Pln}_{\mathrm{arch}},\mathrm{Pln}_{\mathrm{dep}},\mathrm{Pln}_{\mathrm{config}}\}.
$

The analysis agent $\mathcal{A}_\mathrm{analysis}$ subsequently generates detailed implementation analysis for each experimental component specified in the generated plan, such as \texttt{dataloader.py}, \texttt{trainer.py}, etc. Together, the analysis files are denoted as $\mathrm{Analysis}$.

Based on these analyses, the coding agent $\mathcal{A}_\mathrm{code}$ generates the codebase $C$ required for reproduction, which is then executed by the execution agent $\mathcal{A}_\mathrm{exec}$ to produce logs and results $L$. The final workspace comprises: $
E=\{P,\mathrm{Pln},\mathrm{Analysis},C,L\},
$
satisfying the requirement of a complete reproduction $R$. Collectively, these specialised agents constitute the foundation of our framework. %\red{Collectively, these specialised agents constitute the core \textit{Research Agent System} of our framework.}

\begin{algorithm}
\small
\caption{Algorithm for Function \texttt{invoke}}\label{alg}
\begin{algorithmic}[1]
\Require Research Agent $\mathcal{A}_t=\{\mathrm{LLM},\mathrm{Mem}_t,T_t,E,K\}$; Instruction Prompt $p$;
\Ensure Ending Report $\mathrm{Report}$; Updated Workspace $E$;
%\Function{\texttt{invoke}}{$\mathcal{A}_t, p$}
    \State $\mathrm{Mem}_t\gets\mathrm{Mem}_t\cup\{p\}$.
    \For{$i\gets1$ to $K$}
        \State $\mathrm{Reasoning,Action}\gets \mathrm{LLM}(\mathrm{Mem}_t, T_t)$.
            \State $\mathrm{Mem}_t\gets\mathrm{Mem}_t\cup\{(\mathrm{Reasoning,Action})\}$.
        \If{$\mathrm{Action}\in T_t$}
            \State $\mathrm{Result},E\gets\texttt{system.call(}\mathrm{Action}\texttt{)}$.
            \State $\mathrm{Mem}_t\gets\mathrm{Mem}_t\cup\{ \mathrm{Result}\}$.
        \ElsIf{$\mathrm{Action}$ \textbf{matches} $\texttt{"(}\mathcal{A}_\mathrm{sub},p_\mathrm{sub}\texttt{)"}$ \\ \textbf{and}\ $\mathcal{A}_\mathrm{sub}\in\mathrm{Sub}(\mathcal{A}_t)$}
            \State $\mathrm{Result},E\gets$\Call{\texttt{invoke}}{$\mathcal{A}_\mathrm{sub}$, $p_\mathrm{sub}$}.
            \State $\mathrm{Mem}_t\gets\mathrm{Mem}_t\cup\{ \mathrm{Result}\}$.
        \ElsIf{$\mathrm{Action}$ \textbf{matches} $ \texttt{"end\_task(}\mathrm{Report}\texttt{)"}$}
            \State \textbf{return} $\mathrm{Report},E$.
        \EndIf
    \EndFor
    \State $\mathrm{Report}\gets\mathrm{Reasoning}$
    \State \textbf{return} $\mathrm{Report},E$.
%\EndFunction
\end{algorithmic}
\end{algorithm}

\subsection{Hierarchical Orchestration}
 
\ourwork is built on the principle that  reliable research experiment reproduction requires continuous supervision rather than one-time task decomposition. \textit{Manager agents} are therefore granted both global visibility into the shared workspace and authority to actively inspect intermediate artefacts, diagnose failures, and re-invoke subordinate agents with corrective instructions.
%To coordinate the overall paper reproduction workflow, 
We introduce two manager agents into the framework. The first one is the \textit{planning manager} $\mathcal{A}_\mathrm{plan}$, which oversees the entire planning phase by managing all sub-phase agents:

\begin{equation}
\mathrm{Sub}(\mathcal{A}_{\mathrm{plan}})=\{\mathcal{A}_\mathrm{overall},\mathcal{A}_\mathrm{arch},\mathcal{A}_\mathrm{dep},\mathcal{A}_\mathrm{config}\}.
\end{equation}

The second, and the most important, is the \textit{global manager}, which supervises the overall reproduction process at the global phase level, managing the planning agent, analysis agent, coding agent, and execution agent:

\begin{equation}
\mathrm{Sub}(\mathcal{A}_{\mathrm{global}})=\{\mathcal{A}_\mathrm{plan},\mathcal{A}_\mathrm{analysis},\mathcal{A}_\mathrm{code},\mathcal{A}_\mathrm{exec}\}.
\end{equation}

Like the specialised agents, the manager agents are also initialised with the model $\mathrm{LLM}$, their own initial context $\mathrm{Mem}_{t}$, file system toolset $T$, shared environment $E$, the maximum iteration step $K$. Crucially, manager agents differ in that they are granted supervisory authority over the workflow: depending on the current stage of the reproduction process, manager agents can invoke the appropriate subordinate agents with an instruction prompt $p$:

\begin{equation}
\texttt{invoke(}\mathcal{A}_\mathrm{sub},p\texttt{)},\mathcal{A}_\mathrm{sub}\in\mathrm{Sub}(\mathcal{A}_t).
\end{equation}

When an invoked agent is a manager agent, it delegates tasks to the appropriate sub-agents through further invocations. In contrast, a specialised agent directly generates or refines the required artefacts, thereby advancing the reproduction process. The entire reproduction workflow is initiated by explicitly invoking the global manager with an initial prompt $p_0$, i.e., $\texttt{invoke}(\mathcal{A}_\mathrm{global}, p_0)$, which serves as the system entry point. Algorithm~\ref{alg} presents a detailed description of the core \texttt{invoke} function.

After each invocation, the manager agents will inspect the outputs in the workspace via the file system tools $T_\mathrm{file}$. If results are incomplete, lack details, or exhibit signs of hallucination, the manager will instruct the corresponding subordinate agents to re-execute their tasks. This inspection step enables explicit detection and correction of intermediate errors.
 
%\sa{In preliminary experiments, analysis agents frequently hallucinate preprocessing steps or omit critical implementation details described in the paper. In non-hierarchical systems, such errors propagate to the coding and execution stages, resulting in silent failures or incorrect results. In \ourwork, the global manager inspects the workspace, detects missing or inconsistent artefacts, and re-invokes the responsible agent with corrective context, preventing downstream failure.}

Beyond quality control, manager agents also facilitate coordination across subordinate agents. When an agent encounters errors caused by another agent’s outputs, it will report the issue to the manager, who then instructs the responsible agent for correction. This supervision ensures consistency and robust execution of the full reproduction process composed of multiple phases.
%in multi-stage reproduction workflows throughout the full reproduction process.

This design establishes a tree-structured hierarchical orchestration architecture that ensures steady progress, and explicit supervision over intermediate artefacts leading to high-quality outcomes throughout the full reproduction workflow. The initial context prompts for all agents are given in Appendix~\ref{app:agent_prompts}. %and the full prompts will be released in the repository.

\begin{table}[t]
    \centering
    \small
    \begin{adjustbox}{max width=\columnwidth}
    \begin{tabular}{lcc}
        \toprule
        \textbf{Framework} & \textbf{Model} & \textbf{Score (\%)} \\
        \midrule
        %BasicAgent~\cite{starace2025paperbenchevaluatingaisability} & \hyphentt{o3-mini-high} & 6.4 \\
        %\noalign{\vspace{2pt}} 
        %\hdashline
        %\noalign{\vspace{2pt}}
        %\multirow{2}{*}{IterativeAgent~\cite{starace2025paperbenchevaluatingaisability}} & \hyphentt{o3-mini-high} & 17.3 \\
        %& \hyphentt{o1-high} & 43.4 \\
        %\midrule
        
        %\multirow{2}{*}{PaperCoder~\cite{seo2025paper2codeautomatingcodegeneration}} & \hyphentt{o3-mini-high} & 45.1 \\
        AutoReproduce & \hyphentt{o3-mini} & 48.5 \\
        PaperCoder & \multirow{2}{*}{\hyphentt{Claude-Sonnet}} & 51.1 \\
        \hh{\ourwork \textbf{(Ours)}} & & \hh{\textbf{64.1}} \\
        %~\textit{w/ Diagram Understanding}~\cite{zhao2025autoreproduceautomaticaiexperiment} & & 49.6 \\
        \midrule
        PaperCoder* & \multirow{3}{*} {\hyphentt{Qwen3-Coder-480B}} & 36.9 \\
        AutoReproduce* &  & 31.8 \\
        \ourwork \textbf{(Ours)} &  & 45.7 \\
        \noalign{\vspace{2pt}} 
        \hdashline
        \noalign{\vspace{4pt}}
        PaperCoder* & \multirow{5}{*}{\hyphentt{-Terminus}} & 40.8 \\
        AutoReproduce* & \hyphentt{DeepSeek-v3.1} & 41.5 \\
        \ourwork \textbf{(Ours)} & & \textbf{57.4} \\
        \bottomrule
        %\hyphentt{Claude-3.5-Sonnet}
        %\hyphentt{o3-mini-high}
        %~\cite{seo2025paper2codeautomatingcodegeneration}
        %~\cite{zhao2025autoreproduceautomaticaiexperiment}
    \end{tabular}
    \end{adjustbox}
    \vspace{-0.5em}
    \caption{Main results on the PaperBench–CodeDev subset. An asterisk (*) indicates results reproduced by us. 
    %For simplicity, \hyphentt{Claude} denotes \hyphentt{Claude-3.5-Sonnet}, \hyphentt{o3-mini} denotes \hyphentt{o3-mini-high}, \hyphentt{Qwen3} denotes \hyphentt{Qwen3-Coder-480B}, and \hyphentt{DeepSeek-v3.1} denotes \hyphentt{DeepSeek-v3.1-Terminus}. 
    All evaluations in this table are conducted using \hyphentt{o3-mini-high} as the evaluator.}
    \label{tab:sota_results}
    %\vspace{-1.5em}
\end{table}

\section{Experimental Setup}

\begin{table*}[h!]
    \small
    \centering
    \begin{adjustbox}{max width=0.95\textwidth}
    \begin{tabular}{lccccccc}
        \toprule
        \multirow{3}{*}{\textbf{Framework}} 
        & \multicolumn{4}{c}{\textbf{PaperBench-CodeDev (\%)}} 
        & \multicolumn{2}{c}{\textbf{PaperBench (\%)}} \\
        \cmidrule(lr){2-5}\cmidrule(lr){6-7}
        & \multicolumn{2}{c}{\hyphentt{o3-mini-high}}
        & \multicolumn{2}{c}{\hyphentt{ChatGPT-4o-mini}}
        & \multicolumn{2}{c}{\hyphentt{ChatGPT-4o-mini}} \\
        \cmidrule(lr){2-3}\cmidrule(lr){4-5}\cmidrule(lr){6-7}
        & \hyphentt{Qwen3}
        & \hyphentt{DeepSeek-v3.1}
        & \hyphentt{Qwen3}
        & \hyphentt{DeepSeek-v3.1}
        & \hyphentt{Qwen3}
        & \hyphentt{DeepSeek-v3.1} \\
        \midrule
        PaperCoder 
        & 36.9 & 40.8
        & 26.2 & 34.9  
        & 11.5 & 16.9 \\
        AutoReproduce 
        & 31.7 & 41.5 
        & 26.3 & 35.8  
        & 11.2 & 17.1 \\
        \midrule
        Specialised Agents 
        & 37.5 & 47.5
        & 29.9 & 37.9  
        & 15.1 & 18.7 \\
        \textit{+Refinement} 
        & 40.5 & 51.2
        & 31.8 & 39.6  
        & 18.2 & 19.5 \\
        \ourwork (\textbf{Ours}) 
        & \textbf{45.7} & \textbf{57.4}
        & \textbf{34.3} & \textbf{43.1} 
        & \textbf{19.1} & \textbf{21.5} \\
        \bottomrule
    \end{tabular}
    \end{adjustbox}
    \vspace{-0.5em}
    \caption{Extended robustness validation and ablation study on PaperBench-CodeDev and the full PaperBench benchmark. PaperBench-CodeDev results are evaluated using both \hyphentt{ChatGPT-4o-mini} and \hyphentt{o3-mini-high}, while full PaperBench results are evaluated using \hyphentt{ChatGPT-4o-mini} only.}
    \label{tab:ext_result}
\end{table*}

\noindent\textbf{Backbone Models.}~~
%\hh{Our empirical results exhibit a significant performance drop when applying prior frameworks like PaperCoder to open-source LLMs, raising concerns about their robustness beyond proprietary backbones. To address this issue, and also to improve scientific reproducibility and transparency as well as reduce costs, 
%evaluate the scalability of our framework, 
We mainly implement our framework based on two open-source LLMs:
(1) \hyphentt{Qwen3-Coder-480B}~\cite{yang2025qwen3technicalreport}, an MoE model designed for agentic coding and tool use, making it well-suited for experiment reproduction tasks; and (2) \hyphentt{DeepSeek-v3.1-Terminus}~\cite{deepseekai2025deepseekv3technicalreport}, which is optimised for high-performance agentic behaviour in both search and code generation. Using two distinct backbones allows us to disentangle architectural contributions from model-specific effects.
For simplicity, we will also name them as \hyphentt{Qwen3} and \hyphentt{DeepSeek-v3.1}, respectively.

\hh{To further assess generalisability with proprietary models, we additionally employ \hyphentt{Claude-Sonnet} as the backbone in the PaperBench-CodeDev experiment.}

\noindent\textbf{Benchmarks.}~~To evaluate the effectiveness of our framework, we conduct experiments on two experiment reproduction benchmarks. First, \textbf{PaperBench}~\cite{starace2025paperbenchevaluatingaisability} comprises 20 machine learning papers from ICML 2024, each accompanied by a manually constructed evaluation rubric. Second, \textbf{Paper2Code}~\cite{seo2025paper2codeautomatingcodegeneration} contains 90 papers collected from ICML 2024, NeurIPS 2024, and ICLR 2024. These benchmarks jointly assess end-to-end reproduction fidelity across planning, implementation, and execution stages.

\noindent\textbf{Evaluation Protocol.}~~For PaperBench, we adopt the original tree-structured evaluation rubrics provided with each paper, which assess generated repositories along three categories of requirements: code development, execution, and result matching, using percentage scores to indicate comprehensiveness of the reproduction. Following previous work~\cite{seo2025paper2codeautomatingcodegeneration,zhao2025autoreproduceautomaticaiexperiment}, we first evaluate our framework on PaperBench-CodeDev subset with \hyphentt{o3-mini-high}~\cite{ZhangOpenAIOS} as the evaluator to showcase the overall performance. We further use \hyphentt{ChatGPT-4o-mini}~\cite{openai2024gpt4ocard} as an alternative evaluator to assess robustness across evaluators, first validating consistency trends on the PaperBench–CodeDev subset and then extending to the full PaperBench benchmark owing to its cost-efficiency for large-scale evaluation.
%[HH: is this justification reasonable?]} -- JY: good
%We evaluate the proposed approach on both the full benchmark and the Code-Dev subset, which exclusively scores the code development parts of the rubric. Due to budget constraints, we primarily employ \hyphentt{ChatGPT-4o-mini}~\cite{openai2024gpt4ocard} as the evaluator. All methods are evaluated under identical protocols to ensure fair comparison. The evaluator is only provided with the final generated repository and does not have access to internal agent states or intermediate reasoning traces. In addition, to ensure comparability with previous state-of-the-art results, we also report evaluation scores obtained using the DeepSeek model with \hyphentt{o3-mini-high}~\cite{ZhangOpenAIOS} as the evaluator.

%TODO: Change the naming of Specialised agents. 

For the Paper2Code benchmark, evaluation is conducted on a 1–5 scoring scale under two settings: \textit{reference-free} and \textit{reference-based}. The reference-free metric directly evaluates the generated code repository against the paper, whereas the reference-based metric additionally provides the gold-standard repository in the prompt. 

%\vspace{-0.5em}
\begin{figure}[t]
\centering
\tcbset{
    colback=gray!5!white,
    width=0.95\columnwidth,
    fontupper=\small,
    left=1pt,
    right=1pt
}
\begin{tcolorbox}[]
- README.md \\
- config.yaml \\
- datasets/ \\
\tcbtab- data\_loader.py \\
- code/ \\
\tcbtab- model.py \\
- main.py 
\end{tcolorbox}
\vspace{-1em}
\caption{An example of the repository structure}

\vspace{-1.5em}
\label{fig:repo_struct}
\end{figure}

However, our initial studies reveal that reference-free scores based on the original protocol by Paper2Code are systematically overestimated, making them unreliable. To remedy this, we introduce \textbf{Paper2Code-Extra (P2C-Ex)}, a refined reference-free evaluation protocol that enhances the evaluator prompt with repository-level context and additional instructions. Specifically, the prompt includes (i) the total file count in the repository, (ii) a hierarchical repository structure illustration, and (iii) improved instructions to mitigate hallucination. 
The full prompts are presented in Appendix~\ref{app:p2c_prompts}.
An example of repository structure is displayed in Figure~\ref{fig:repo_struct}. %\hh{[HH: Can we use examples in Appendix~\ref{app:case_examples} Figure~\ref{fig:repo_examples} to replace the example here?]}
%\red{[CL: this is actually a selling point of our contribution; should have provided more details about Paper2Code-Extra than simply pointing readers to the appendix]}
%Further details are provided in \S\ref{sec:p2c}. 
To quantify the consistency between different evaluation protocols, we report the Pearson correlation coefficient ($r$) between the reference-based setting and each reference-free setting. All Paper2Code evaluations are conducted using \hyphentt{o3-mini-high} as the evaluator.

\noindent\textbf{Baselines.}~~We evaluate our framework against two baselines for experiment reproduction. 
\textbf{PaperCoder}~\cite{seo2025paper2codeautomatingcodegeneration} employs a sequential multi-stage pipeline with a dedicated agent for each stage to generate codebases from papers. \textbf{AutoReproduce}~\cite{zhao2025autoreproduceautomaticaiexperiment} is a multi-agent system with search and coding agents that improve reproduction quality by leveraging information from related papers. For a fair comparison, we reproduced both baselines using the \hyphentt{Qwen3} and \hyphentt{DeepSeek-v3.1} backbones. 
%The BasicAgent and IterativeAgent are single-agent baselines from the original PaperBench paper. We only gather their reported performance for state-of-the-art comparison.

For the Paper2Code benchmark, we additionally include three naïve baselines to demonstrate the intrinsic limitation of the original reference-free evaluation: (i) an empty repository, (ii) repositories containing only configuration or documentation files (e.g., \texttt{.md} or \texttt{.yaml}) extracted from the gold reference repository, and (iii) the gold reference repositories themselves. 
%\red{[CL: briefly explain why need these three naive baselines]}

\noindent\textbf{Environment.}~~All experiments are conducted with eight NVIDIA H200. We use \texttt{smolagents} \cite{smolagents} as our agent scaffold.

\section{Experimental Results}
%TODO: baseline decriptions.
\begin{table*}[h!]
    \centering
    \small
    \begin{adjustbox}{max width=0.95\textwidth}
    \begin{tabular}{lcccccc}
        \toprule
        & \multirow{2}{*}{\textbf{Model}} & \multicolumn{5}{c}{\textbf{Score (5-point scale)}} \\
        \cmidrule(lr){3-7}
        & & Ref-Based & Ref-Free & \textit{+Count} & \textit{+Structure} & P2C-Ex\\
        \midrule
        Empty Repo & \multirow{3}{*}{-}& 1.57 & 3.89 & 1.0 & 1.0 & 1.0 \\
        Config Only & & 1.62 & 4.67 & 3.96 & 3.41 & 2.68 \\
        Gold Repo & & - & 4.75 & 4.82 & 4.84 & 4.80 \\
        \midrule
        PaperCoder & \texttt{o3-mini-high} & 3.66 & 4.55 & - & - & - \\
        \midrule
        PaperCoder* & \multirow{3}{*}{\hyphentt{Qwen3-Coder-480B}} & 1.83 & 2.97 & 2.26 & 1.68 &1.58 \\
        AutoReproduce* & & 2.45 & 3.03 & 2.74 & 2.31 & 2.09 \\
        \ourwork \textbf{(Ours)} & & \textbf{2.45} & \textbf{3.15} & \textbf{3.03} & \textbf{2.98} & \textbf{2.96} \\
        \noalign{\vspace{0.4ex}} 
        \hdashline
        \noalign{\vspace{0.65ex}} 
        PaperCoder* & \multirow{3}{*}{\hyphentt{DeepSeek-v3.1-Terminus}} & 2.25 & 4.38 & 2.74 & 2.31 & 2.09 \\
        AutoReproduce* &  & 2.51 & 3.59 & 3.60 & 3.58 & 3.59\\
        \ourwork \textbf{(Ours)} & & \textbf{3.64} & \textbf{4.42} & \textbf{3.94} & \textbf{3.88} & \textbf{3.86} \\
        \midrule
        Pearson Correlation $r$ & - & - & 0.423 & 0.724 & 0.797 & \textbf{0.862}\\
        \bottomrule
    \end{tabular}
    \end{adjustbox}
    \vspace{-0.5em}
    \caption{Evaluation results on the Paper2Code benchmark using reference-based (Ref-Based) and reference-free evaluation strategies. For reference-free evaluations, we include the original reference-free prompt of Paper2Code (Ref-Free), the step-by-step enhancements for ablation study (\textit{+Count}, \textit{+Structure}), and our final revised version (P2C-Ex).  Pearson correlation $r$ measures the correlation between the reference-based scores and the scores by each of the reference-free evaluations. All evaluations are conducted using \hyphentt{o3-mini-high} as the evaluator.}
    \label{tab:p2c_result}
    %\vspace{-1.5em}
\end{table*}
\subsection{Overall Results}

Table~\ref{tab:sota_results} reports the main results on the PaperBench-CodeDev subset. Prior frameworks exhibit substantial performance degradation when transferred from proprietary to open-source backbones. For instance, PaperCoder drops from 51.1\% on \hyphentt{Claude-Sonnet} to 40.8\% on \hyphentt{DeepSeek-v3.1}, and further to 36.9\% on \hyphentt{Qwen3}, corresponding to a relative decrease of over 20\%. \hh{In contrast, our framework establishes a new state-of-the-art with 64.1\% on \hyphentt{Claude-Sonnet}, while maintaining strong performance on open-source models. Specifically, it achieves 57.4\% on \hyphentt{DeepSeek-v3.1}, surpassing the previous best result of 51.1\% obtained with proprietary backbones. Our method also consistently outperforms prior approaches on \hyphentt{Qwen3}.}
%Table~\ref{tab:sota_results} presents the main results on the PaperBench–CodeDev subset. As shown, prior frameworks experience a substantial performance degradation when transferred from proprietary to open-source backbones. For example, PaperCoder drops from 51.1\% on \hyphentt{Claude-3.5-Sonnet} to 40.8\% on \hyphentt{DeepSeek-v3.1} and further to 36.9\% on \hyphentt{Qwen3}, corresponding to over a 20\% relative decrease. In contrast, our framework performs strongly even with open-sourced models, achieving a new state-of-the-art of 57.4\% on \hyphentt{DeepSeek-v3.1}, surpassing even the best proprietary model result.  Our results on \hyphentt{Qwen3} also outperforms prior methods. 
The resulting 25.4\% relative improvement over the previous SOTA score indicates that our approach more effectively leverages the latent capabilities of open-source LLMs, enabling performance that matches or exceeds levels previously attainable only with proprietary models.

\hh{Table \ref{tab:ext_result} presents more detailed results on PaperBench with the two open-source models.}Across all settings, \hyphentt{DeepSeek-v3.1} consistently outperforms \hyphentt{Qwen3}, suggesting stronger coding proficiency and research-oriented reasoning that is particularly beneficial for agent-based reproduction systems. Importantly, the relative gains achieved by our method are preserved across both backbones, with improvements of 38.3\% on \hyphentt{DeepSeek-v3.1} and 23.8\% on \hyphentt{Qwen3} over the strongest corresponding baselines. This consistency demonstrates the robustness and generalisability of our framework across model architectures.

The results on the Paper2Code benchmark are presented in Table~\ref{tab:p2c_result}. Across all evaluation settings and backbone models, our framework consistently achieves the highest scores, with \hyphentt{DeepSeek-v3.1} outperforming \hyphentt{Qwen3}, consistent with trends observed in previous experiments. Beyond absolute performance, the markedly different magnitudes of absolute improvement under the reference-based and reference-free metrics (1.39 vs. 0.04) motivate a closer examination of the evaluation protocol. %, pointing to a critical limitation of the original reference-free metric.

\subsection{Paper2Code-Extra Meta-Evaluation}
\label{sec:p2c}

Closer inspection reveals a critical limitation of the original reference-free metric: it systematically overestimates repository quality. In particular, repositories with few or no executable files can receive disproportionately high scores. To illustrate this issue, we conduct extra experiments on naïve baselines, as shown in Table~\ref{tab:p2c_result}: (i) the \textit{empty repository} attains a score of 3.89 under the reference-free metric, exceeding all \hyphentt{Qwen3}-based results, and (ii) \textit{configuration-only} repositories containing no executable code can score 4.67, outperforming all model-generated codebases. These anomalies indicate that the reference-free metric is intrinsically unreliable. Further inspection attributes this issue to evaluator hallucination, whereby even strong evaluators such as \hyphentt{o3-mini} infer non-existent code files or misinterpret documentation as implemented code in the empty or config-only repositories.

To address this issue, we propose \textbf{Paper2Code-Extra (P2C-Ex)}, a revised reference-free evaluation paradigm that augments the prompt with explicit repository-level information and refined instructions. %\hh{[HH: A bit repetitive with the Experimental Setup?]} 
As illustrated in the meta-evaluation results, incorporating the information substantially improves alignment with the reference-based metric: the Pearson correlation coefficient $r$ increases from 0.423 to 0.862. Consistent trends are observed in repository-level regression analyses (Appendix~\ref{app:linreg}), where $r$ improves from 0.42 to 0.83. Moreover, the prevalence of severely overrated repositories under the original protocol (Figure~\ref{fig:ref_free}) is remarkably mitigated with P2C-Ex (Figure~\ref{fig:p2c_ex}). 

Ablation results for P2C-Ex further demonstrate that file count, repository structure, and revised prompt instructions each contribute to mitigating evaluator hallucination. Together, these results validate the effectiveness of P2C-Ex and its improved alignment with the reference-based metric, highlighting the importance of structurally informed prompts for reliable reference-free evaluation. 

\begin{table}[t]
    \centering
    \small
    \begin{adjustbox}{max width=\columnwidth}
    \begin{tabular}{lc}
        \toprule
        \textbf{Metric} & \textbf{Pearson Correlation} \\
        \midrule
        Inter-Annotator & 0.82\\
        Ref-Based & 0.77\\
        \midrule
        Ref-Based & 0.50\\
        \textit{+Count} & 0.59\\
        \textit{+Structure} & 0.64\\
        P2C-Ex(\textbf{Ours}) & \textbf{0.72}\\
        \bottomrule
        %\hyphentt{Claude-3.5-Sonnet}
        %\hyphentt{o3-mini-high}
        %~\cite{seo2025paper2codeautomatingcodegeneration}
        %~\cite{zhao2025autoreproduceautomaticaiexperiment}
    \end{tabular}
    \end{adjustbox}
    \vspace{-0.5em}
    \caption{Pearson correlation between different evaluation metrics and human expert annotations. "Inter-Annotator" denotes the average Pearson correlation among three experts.}
    \label{tab:correlation}
    %\vspace{-1.5em}
\end{table}

\hh{To further validate the alignment between our reference-free evaluation metric and human judgment, we collect expert annotations from three evaluators on a subset of 30 papers from ICLR 2024 within the Paper2Code benchmark.  Table~\ref{tab:correlation} reports the Pearson correlation between each metric and the expert annotations. The results demonstrate that P2C-Ex achieves the highest correlation with human evaluations, indicating strong reliability. Moreover, the correlation consistently improves as additional information is incorporated into the prompt.}

\subsection{Ablation Study}
To rigorously evaluate the contributions of individual components, we further perform an ablation study on %the \hyphentt{DeepSeek-v3.1} backbone
PaperBench, with results reported in Table~\ref{tab:ext_result}. Our framework is decomposed into three key elements: (i) a single-step research agent with specialised agents, (ii) adding intra-agent iterative refinement with $K$ steps, and (iii) introducing inter-agent hierarchical management via manager agents, forming \ourwork.

The specialised-agent pipeline alone outperforms prior research agent frameworks, demonstrating that finer-grained decomposition of responsibilities for phase-specific agents leads to more effective experiment reproduction than monolithic designs. Furthermore, incorporating iterative self-refinement yields substantial additional gains, enabling performance that marginally surpasses previous state-of-the-art baselines. This highlights the ability of specialised agents to identify and correct errors through environment interaction. Finally, hierarchical management fully harnesses the capabilities of the backbone models, delivering the strongest performance across all settings. Moreover, hierarchical supervision alone contributes approximately a 10\% increase, underscoring its critical role in mitigating error propagation and coordinating long-horizon reproduction processes. Collectively, these results further validate the effectiveness of our hierarchical multi-agent system when all components are integrated.

\subsection{Extended Results Across Evaluators}
\label{sec:ablation_study}
%%\hh{[HH: Better Section Title?]}
%
%Table~\ref{tab:ext_result} reports an extended robustness evaluation on both the PaperBench–CodeDev subset and the full PaperBench benchmark using \hyphentt{ChatGPT-4o-mini} as the evaluator.

Table~\ref{tab:ext_result} reports additional results on PaperBench using a different evaluator, \hyphentt{ChatGPT-4o-mini}. The findings are consistent with those obtained using \hyphentt{o3-mini-high}.
\ourwork outperforms all the baselines across all settings, achieving relative improvements of 30.4\% on \hyphentt{Qwen3} and 20.4\% on \hyphentt{DeepSeek-v3.1}, demonstrating robustness to different evaluators. Notably, these advantages not only persist but become more pronounced on the full PaperBench benchmark, which additionally evaluates execution outcomes and is therefore substantially more challenging, as reflected by lower absolute scores. In this setting, our framework attains larger relative gains of 66.1\% on \hyphentt{Qwen3} and 25.7\% on \hyphentt{DeepSeek-v3.1}, underscoring the critical role of the \textit{execution} phase in end-to-end reproduction.
%Taken together, these consistent gains indicate that our framework more effectively translates the latent capabilities of LLMs into reliable code development and experimental reproduction, rather than relying on model scale alone.

\section{Qualitative Analysis}

\subsection{Case Study}
Appendix~\ref{app:case_examples} compares PaperCoder and \ourwork on reproducing the same PaperBench paper. As shown in Figure~\ref{fig:repo_examples}, PaperCoder produces a flat repository, %by sequentially prompting and externally extracting code snippets
whereas \ourwork generates a well-organised codebase with clearly separated subdirectories. This improvement can be attributed to our specialised agents’ ability to directly interact with the shared workspace via file-system tools, enabling structured artefact management. In addition, execution agents create auxiliary test files to detect runtime errors, which are reported to manager agents and used to guide subsequent code revisions.

A closer examination %of the reproduction process 
further highlights the role of hierarchical supervision. As shown in Figure~\ref{fig:manager_prompt}, the planning manager detects insufficient detail in the initial implementation plan and re-invokes the overall planning agent for refinement, resulting in a substantially more detailed implementation roadmap than PaperCoder (see Figure~\ref{fig:plan_examples}). For example, our plan explicitly specifies architectural components such as the structure of the encoder and decoder in the Deep Generative Model, which is absent in PaperCoder’s plan. This increased planning fidelity prevents errors introduced at early stages from propagating and facilitates more effective downstream implementation.

These qualitative differences are eventually reflected in the reproduction scores on this paper: \ourwork achieves 66.6\%, nearly doubling PaperCoder’s 33.8\%. Overall, this case study highlights the proficiency of our specialised agents and the importance of manager agents in controlling reproduction quality and mitigating error propagation.

\subsection{\hh{Failure Mode Analysis}}

To better characterise failure modes in our reproduction pipeline, we conduct a detailed manual analysis of representative cases. Our results show that failures predominantly arise during execution rather than code generation, as reflected by a substantial performance drop from approximately 45\% on PaperBench-CodeDev to 20\% on the full PaperBench across all models.

Our analysis further indicates that most papers yield well-structured codebases, while execution introduces the majority of errors. For example, when using \hyphentt{DeepSeek-v3.1} as the backbone to reproduce all papers from both PaperBench and Paper2Code, only 3 out of 110 runs fail at the structured code generation stage, demonstrating strong robustness in code development. In contrast, the predominant failure mode stems from execution errors due to incorrect inter-file dependencies, particularly in complex directory structures. A detailed case study is provided in Appendix~\ref{app:failure_analysis}, where the evaluation score drops from 69.7\% to 11.3\% when execution is included, largely due to import failures. Additional issues include failures in environment setup and in downloading required models or datasets from platforms such as GitHub and Hugging Face. Overall, the execution stage remains the primary bottleneck of current models and frameworks.

%\vspace{5pt}
\section{Conclusion}
%\vspace{5pt}

In this work, we present \ourwork, a hierarchical multi-agent framework for end-to-end experiment reproduction, introducing manager agents to coordinate the multi-stage workflow. Comprehensive experiments illustrate that our framework consistently outperforms prior approaches on experiment reproduction benchmarks with state-of-the-art performance achieved by open-source models, highlighting the benefits of hierarchical supervision and specialised agent collaboration across the system. We also propose a revised evaluation protocol, P2C-Ex, that leverages repository-level information for stronger alignment with reference-based metrics. Moreover, the case study underscores how hierarchical coordination improves the overall quality of reproduction and mitigates error propagation. Collectively, these contributions advance both the methodology and evaluation of automated experiment reproduction, providing practical insights for using LLM-based agents in scientific research.

%In this work, we present \ourwork, a hierarchical multi-agent framework for automated research paper reproduction, introducing manager agents to coordinate complex, multi-stage workflows. Comprehensive experiments illustrate that our framework consistently outperforms prior approaches on paper reproduction benchmarks with new state-of-the-art performance, highlighting the benefits of hierarchical supervision and collaboration of specialised agents across the system. We also reveal a critical limitation of existing reference-free evaluation protocols and propose a revised paradigm, P2C-Ex, that leverages repository-level information for stronger alignment with reference-based metrics. \hh{In addition, we provide qualitative analyses that illustrate how hierarchical coordination improves the overall quality of reproduction.} Collectively, these contributions advance both the methodology and evaluation of automated paper reproduction, providing practical insights for using LLM-based agents in scientific research.

\section*{Limitations}

Due to budget constraints, we do not evaluate all experimental settings with the \hyphentt{o3-mini} model. However, all reported comparisons are conducted under consistent evaluation protocols to ensure fairness across methods. In addition, our method may incur higher time and token costs than prior approaches, stemming from the increased complexity of agent reasoning and tool calling.
This overhead reflects an inherent trade-off for introducing explicit supervision and error correction, and our results indicate that the additional cost yields substantial improvements in reproduction robustness and quality.

\section*{Ethical Considerations}

All data in our work are collected from publicly available sources.
Our primary objective is to support researchers in reproducing prior works, rather than replacing the creative and critical activities in scientific research. By design, the system produces explicit intermediate artefacts and execution logs, enabling human inspection and verification of the reproduction process. The system should be used as a supplementary tool to aid, rather than substitute, methodological rigour and independent research judgment.

%\section*{Acknowledgements}
\bibliography{custom}

@misc{starace2025paperbenchevaluatingaisability,
      title={PaperBench: Evaluating AI's Ability to Replicate AI Research}, 
      author={Giulio Starace and Oliver Jaffe and Dane Sherburn and James Aung and Jun Shern Chan and Leon Maksin and Rachel Dias and Evan Mays and Benjamin Kinsella and Wyatt Thompson and Johannes Heidecke and Amelia Glaese and Tejal Patwardhan},
      year={2025},
      eprint={2504.01848},
      archivePrefix={arXiv},
      primaryClass={cs.AI},
      url={https://arxiv.org/abs/2504.01848}, 
}

@article{james2026rigourate,
  title={RIGOURATE: Quantifying Scientific Exaggeration with Evidence-Aligned Claim Evaluation},
  author={James, Joseph and Xiao, Chenghao and Li, Yucheng and Moosavi, Nafise Sadat and Lin, Chenghua},
  journal={arXiv preprint arXiv:2601.04350},
  year={2026}
}

@article{eger2025transforming,
  title={Transforming science with large language models: A survey on ai-assisted scientific discovery, experimentation, content generation, and evaluation},
  author={Eger, Steffen and Cao, Yong and D'Souza, Jennifer and Geiger, Andreas and Greisinger, Christian and Gross, Stephanie and Hou, Yufang and Krenn, Brigitte and Lauscher, Anne and Li, Yizhi and others},
  journal={arXiv preprint arXiv:2502.05151},
  year={2025}
}

@misc{seo2025paper2codeautomatingcodegeneration,
      title={Paper2Code: Automating Code Generation from Scientific Papers in Machine Learning}, 
      author={Minju Seo and Jinheon Baek and Seongyun Lee and Sung Ju Hwang},
      year={2025},
      eprint={2504.17192},
      archivePrefix={arXiv},
      primaryClass={cs.CL},
      url={https://arxiv.org/abs/2504.17192}, 
}

@misc{zhao2025autoreproduceautomaticaiexperiment,
      title={AutoReproduce: Automatic AI Experiment Reproduction with Paper Lineage}, 
      author={Xuanle Zhao and Zilin Sang and Yuxuan Li and Qi Shi and Weilun Zhao and Shuo Wang and Duzhen Zhang and Xu Han and Zhiyuan Liu and Maosong Sun},
      year={2025},
      eprint={2505.20662},
      archivePrefix={arXiv},
      primaryClass={cs.AI},
      url={https://arxiv.org/abs/2505.20662}, 
}

@misc{yang2025qwen3technicalreport,
      title={Qwen3 Technical Report}, 
      author={An Yang and Anfeng Li and Baosong Yang and Beichen Zhang and Binyuan Hui and Bo Zheng and Bowen Yu et al.},
      year={2025},
      eprint={2505.09388},
      archivePrefix={arXiv},
      primaryClass={cs.CL},
      url={https://arxiv.org/abs/2505.09388}, 
}

@misc{deepseekai2025deepseekv3technicalreport,
      title={DeepSeek-V3 Technical Report}, 
      author={DeepSeek-AI and Aixin Liu and Bei Feng and Bing Xue and Bingxuan Wang and Bochao Wu and Chengda Lu and Chenggang Zhao and Chengqi Deng and Chenyu Zhang and Chong Ruan and Damai Dai and Daya Guo and Dejian Yang and Deli Chen and Dongjie Ji and Erhang Li and Fangyun Lin and Fucong Dai and Fuli Luo and Guangbo Hao and Guanting Chen and Guowei Li and H. Zhang and Han Bao and Hanwei Xu and Haocheng Wang and Haowei Zhang and Honghui Ding and Huajian Xin and Huazuo Gao and Hui Li and Hui Qu and J. L. Cai and Jian Liang and Jianzhong Guo and Jiaqi Ni and Jiashi Li and Jiawei Wang and Jin Chen and Jingchang Chen and Jingyang Yuan and Junjie Qiu and Junlong Li and Junxiao Song and Kai Dong and Kai Hu and Kaige Gao and Kang Guan and Kexin Huang and Kuai Yu and Lean Wang and Lecong Zhang and Lei Xu and Leyi Xia and Liang Zhao and Litong Wang and Liyue Zhang and Meng Li and Miaojun Wang and Mingchuan Zhang and Minghua Zhang and Minghui Tang and Mingming Li and Ning Tian and Panpan Huang and Peiyi Wang and Peng Zhang and Qiancheng Wang and Qihao Zhu and Qinyu Chen and Qiushi Du et al.},
      year={2025},
      eprint={2412.19437},
      archivePrefix={arXiv},
      primaryClass={cs.CL},
      url={https://arxiv.org/abs/2412.19437}, 
}

@misc{openai2024gpt4ocard,
      title={GPT-4o System Card}, 
      author={OpenAI},
      year={2024},
      eprint={2410.21276},
      archivePrefix={arXiv},
      primaryClass={cs.CL},
      url={https://arxiv.org/abs/2410.21276}, 
}

@inproceedings{ZhangOpenAIOS,
  title={OpenAI o3-mini System Card},
  author={Brian Zhang and Eric Mitchell and Hongyu Ren and Kevin Lu and Max Schwarzer and Michelle Pokrass and Shengjia Zhao and Ted Sanders and Adam Tauman Kalai and Alexandre Passos and Benjamin Sokolowsky and Elaine Ya Le and Erik Ritter and Hao Sheng and Hanson Wang and Ilya Kostrikov and James Lee and Johannes Ferstad and Michael Lampe and Prashanth Radhakrishnan and Sean Fitzgerald and S{\'e}bastien Bubeck and Yann Dubois and Yu Bai and Andy Applebaum and Elizabeth Proehl and Evan Mays and Joel Parish and Kevin Liu and Leon Maksin and Leyton Ho and Miles Wang and Michele Wang and Olivia Watkins and Patrick Chao and Samuel Miserendino and Tejal Patwardhan and Antonia Woodford and Beth Hoover and Jake Brill and Kelly Stirman and Neel Ajjarapu and Nick Turley and Nikunj Handa and Olivier Godement and Akshay Nathan and Alyssa Huang and Andy Wang and Ankit Gohel and Ben Eggers and Brian Yu and Bryan Ashley and Chengdu Huang and Davin Bogan and Emily Sokolova and Eric Horacek and Felipe Petroski Such and Jonah Cohen and Joshua Gross and Justin Becker and Kan Wu and Larry Lv and Lee Byron and Manoli Liodakis and Max Johnson and Mike Trpcic and Murat Yesildal and Rasmus Rygaard and R. J. Marsan and Rohit Ram-chandani and Rohan Kshirsagar and Sara Conlon and Tony Xia and Siyuan Fu and Srinivas Narayanan and Sulman Choudhry and Tomer Kaftan and Trevor Creech and Andrea Vallone and Andrew Duberstein and Enis Sert and Eric Wallace and Grace Zhao and Irina Kofman and Jieqi Yu and Joaquin Qui{\~n}onero Candela and Made-laine Boyd and Mehmet Ali Yatbaz and Mike McClay and Mingxuan Wang and Sandhini Agarwal and Saachi Jain and Sam Toizer and Santiago Hern{\'a}ndez and Steve Mostovoy and Tao Li and Young Cha and Yunyun Wang and Lama Ahmad and Troy Peterson and Carpus Chang and Kristen Ying and Aidan Clark and Dane Stuckey and Jerry Tworek and Jakub W. Pachocki and Johannes Heidecke and Kevin Weil and Liam Fedus and Mark Chen and Sam Altman and Wojciech Zaremba},
  url={https://cdn.openai.com/o3-mini-system-card-feb10.pdf},
  year={2025}
}

@inproceedings{magnusson-etal-2023-reproducibility,
    title = "Reproducibility in {NLP}: What Have We Learned from the Checklist?",
    author = "Magnusson, Ian  and
      Smith, Noah A.  and
      Dodge, Jesse",
    editor = "Rogers, Anna  and
      Boyd-Graber, Jordan  and
      Okazaki, Naoaki",
    booktitle = "Findings of the Association for Computational Linguistics: ACL 2023",
    month = jul,
    year = "2023",
    address = "Toronto, Canada",
    publisher = "Association for Computational Linguistics",
    url = "https://aclanthology.org/2023.findings-acl.809/",
    doi = "10.18653/v1/2023.findings-acl.809",
    pages = "12789--12811"
}

@Misc{smolagents,
  title =        {`smolagents`: a smol library to build great agentic systems.},
  author =       {Aymeric Roucher and Albert Villanova del Moral and Thomas Wolf and Leandro von Werra and Erik Kaunismäki},
  howpublished = {\url{https://github.com/huggingface/smolagents}},
  year =         {2025}
}

@inproceedings{
yao2023react,
title={ReAct: Synergizing Reasoning and Acting in Language Models},
author={Shunyu Yao and Jeffrey Zhao and Dian Yu and Nan Du and Izhak Shafran and Karthik R Narasimhan and Yuan Cao},
booktitle={The Eleventh International Conference on Learning Representations },
year={2023},
url={https://openreview.net/forum?id=WE_vluYUL-X}
}

@misc{yamada2025aiscientistv2workshoplevelautomated,
      title={The AI Scientist-v2: Workshop-Level Automated Scientific Discovery via Agentic Tree Search}, 
      author={Yutaro Yamada and Robert Tjarko Lange and Cong Lu and Shengran Hu and Chris Lu and Jakob Foerster and Jeff Clune and David Ha},
      year={2025},
      eprint={2504.08066},
      archivePrefix={arXiv},
      primaryClass={cs.AI},
      url={https://arxiv.org/abs/2504.08066}, 
}

@article{zochi2025,
  title={Zochi Technical Report},
  author={Intology},
  journal={arXiv},
  year={2025}
}

@inproceedings{li-etal-2025-chain-ideas,
    title = "Chain of Ideas: Revolutionizing Research Via Novel Idea Development with {LLM} Agents",
    author = "Li, Long  and
      Xu, Weiwen  and
      Guo, Jiayan  and
      Zhao, Ruochen  and
      Li, Xingxuan  and
      Yuan, Yuqian  and
      Zhang, Boqiang  and
      Jiang, Yuming  and
      Xin, Yifei  and
      Dang, Ronghao  and
      Rong, Yu  and
      Zhao, Deli  and
      Feng, Tian  and
      Bing, Lidong",
    editor = "Christodoulopoulos, Christos  and
      Chakraborty, Tanmoy  and
      Rose, Carolyn  and
      Peng, Violet",
    booktitle = "Findings of the Association for Computational Linguistics: EMNLP 2025",
    month = nov,
    year = "2025",
    address = "Suzhou, China",
    publisher = "Association for Computational Linguistics",
    url = "https://aclanthology.org/2025.findings-emnlp.477/",
    doi = "10.18653/v1/2025.findings-emnlp.477",
    pages = "8971--9004",
    ISBN = "979-8-89176-335-7"
}

@inproceedings{james-etal-2024-rigour,
    title = "On the Rigour of Scientific Writing: Criteria, Analysis, and Insights",
    author = "James, Joseph  and
      Xiao, Chenghao  and
      Li, Yucheng  and
      Lin, Chenghua",
    editor = "Al-Onaizan, Yaser  and
      Bansal, Mohit  and
      Chen, Yun-Nung",
    booktitle = "Findings of the Association for Computational Linguistics: EMNLP 2024",
    month = nov,
    year = "2024",
    address = "Miami, Florida, USA",
    publisher = "Association for Computational Linguistics",
    url = "https://aclanthology.org/2024.findings-emnlp.380/",
    doi = "10.18653/v1/2024.findings-emnlp.380",
    pages = "6523--6538"
}

@article{DBLP:journals/corr/abs-2502-14743,
  publtype={informal},
  author={Lijun Sun and Yijun Yang and Qiqi Duan and Yuhui Shi and Chao Lyu and Yu-Cheng Chang and Chin-Teng Lin and Yang Shen},
  title={Multi-Agent Coordination across Diverse Applications: A Survey},
  year={2025},
  month={February},
  cdate={1738368000000},
  journal={CoRR},
  volume={abs/2502.14743},
  url={https://doi.org/10.48550/arXiv.2502.14743}
}

@article{mooretaxonomy,
  title={A Taxonomy of Hierarchical Multi-Agent Systems: Design Patterns, Coordination Mechanisms, and Industrial Applications},
  author={Moore, David},
  journal={Coordination Mechanisms, and Industrial Applications},
  year={2025}
}

@article{DBLP:journals/corr/abs-1901-08492,
  author       = {Sanjeevan Ahilan and
                  Peter Dayan},
  title        = {Feudal Multi-Agent Hierarchies for Cooperative Reinforcement Learning},
  journal      = {CoRR},
  volume       = {abs/1901.08492},
  year         = {2019},
  url          = {http://arxiv.org/abs/1901.08492},
  eprinttype    = {arXiv},
  eprint       = {1901.08492},
  bibsource    = {dblp computer science bibliography, https://dblp.org}
}

@inproceedings{
wu2024autogen,
title={AutoGen: Enabling Next-Gen {LLM} Applications via Multi-Agent Conversations},
author={Qingyun Wu and Gagan Bansal and Jieyu Zhang and Yiran Wu and Beibin Li and Erkang Zhu and Li Jiang and Xiaoyun Zhang and Shaokun Zhang and Jiale Liu and Ahmed Hassan Awadallah and Ryen W White and Doug Burger and Chi Wang},
booktitle={First Conference on Language Modeling},
year={2024},
url={https://openreview.net/forum?id=BAakY1hNKS}
}

@misc{xie2025faraiscientistschanging,
      title={How Far Are AI Scientists from Changing the World?}, 
      author={Qiujie Xie and Yixuan Weng and Minjun Zhu and Fuchen Shen and Shulin Huang and Zhen Lin and Jiahui Zhou and Zilan Mao and Zijie Yang and Linyi Yang and Jian Wu and Yue Zhang},
      year={2025},
      eprint={2507.23276},
      archivePrefix={arXiv},
      primaryClass={cs.AI},
      url={https://arxiv.org/abs/2507.23276}, 
}

@misc{gottweis2025aicoscientist,
      title={Towards an AI co-scientist}, 
      author={Juraj Gottweis and Wei-Hung Weng and Alexander Daryin and Tao Tu and Anil Palepu and Petar Sirkovic and Artiom Myaskovsky and Felix Weissenberger and Keran Rong and Ryutaro Tanno and Khaled Saab and Dan Popovici and Jacob Blum and Fan Zhang and Katherine Chou and Avinatan Hassidim and Burak Gokturk and Amin Vahdat and Pushmeet Kohli and Yossi Matias and Andrew Carroll and Kavita Kulkarni and Nenad Tomasev and Yuan Guan and Vikram Dhillon and Eeshit Dhaval Vaishnav and Byron Lee and Tiago R D Costa and José R Penadés and Gary Peltz and Yunhan Xu and Annalisa Pawlosky and Alan Karthikesalingam and Vivek Natarajan},
      year={2025},
      eprint={2502.18864},
      archivePrefix={arXiv},
      primaryClass={cs.AI},
      url={https://arxiv.org/abs/2502.18864}, 
}

@inproceedings{
si2025can,
title={Can {LLM}s Generate Novel Research Ideas? A Large-Scale Human Study with 100+ {NLP} Researchers},
author={Chenglei Si and Diyi Yang and Tatsunori Hashimoto},
booktitle={The Thirteenth International Conference on Learning Representations},
year={2025},
url={https://openreview.net/forum?id=M23dTGWCZy}
}

@inproceedings{guo2024large,
  title={Large Language Model Based Multi-agents: A Survey of Progress and Challenges},
  author={Guo, Taicheng and Chen, Xiuying and Wang, Yaqi and Chang, Ruidi and Pei, Shichao and Chawla, Nitesh V and Wiest, Olaf and Zhang, Xiangliang},
  booktitle={IJCAI},
  year={2024}
}

@article{li2024survey,
  title={A survey on LLM-based multi-agent systems: workflow, infrastructure, and challenges},
  author={Li, Xinyi and Wang, Sai and Zeng, Siqi and Wu, Yu and Yang, Yi},
  journal={Vicinagearth},
  volume={1},
  number={1},
  pages={9},
  year={2024},
  publisher={Springer}
}

@article{sun2025llm,
  title={LLM-Based Multi-Agent Decision-Making: Challenges and Future Directions},
  author={Sun, Chuanneng and Huang, Songjun and Pompili, Dario},
  journal={IEEE Robotics and Automation Letters},
  year={2025},
  publisher={IEEE}
}

@inproceedings{
xiang2025scireplicatebench,
title={SciReplicate-Bench: Benchmarking {LLM}s in Agent-driven Algorithmic Reproduction from Research Papers},
author={Yanzheng Xiang and Hanqi Yan and Shuyin Ouyang and Lin Gui and Yulan He},
booktitle={Second Conference on Language Modeling},
year={2025},
url={https://openreview.net/forum?id=8LoPjpvWde}
}

@article{DBLP:journals/corr/abs-2408-06292,
  publtype={informal},
  author={Chris Lu and Cong Lu and Robert Tjarko Lange and Jakob N. Foerster and Jeff Clune and David Ha},
  title={The AI Scientist: Towards Fully Automated Open-Ended Scientific Discovery},
  year={2024},
  cdate={1704067200000},
  journal={CoRR},
  volume={abs/2408.06292},
  url={https://doi.org/10.48550/arXiv.2408.06292}
}

@ARTICLE{9849664,
  author={Dehaerne, Enrique and Dey, Bappaditya and Halder, Sandip and De Gendt, Stefan and Meert, Wannes},
  journal={IEEE Access}, 
  title={Code Generation Using Machine Learning: A Systematic Review}, 
  year={2022},
  volume={10},
  number={},
  pages={82434-82455},
  doi={10.1109/ACCESS.2022.3196347}}

@misc{chen2025ai4researchsurveyartificialintelligence,
      title={AI4Research: A Survey of Artificial Intelligence for Scientific Research}, 
      author={Qiguang Chen and Mingda Yang and Libo Qin and Jinhao Liu and Zheng Yan and Jiannan Guan and Dengyun Peng and Yiyan Ji and Hanjing Li and Mengkang Hu and Yimeng Zhang and Yihao Liang and Yuhang Zhou and Jiaqi Wang and Zhi Chen and Wanxiang Che},
      year={2025},
      eprint={2507.01903},
      archivePrefix={arXiv},
      primaryClass={cs.CL},
      url={https://arxiv.org/abs/2507.01903}, 
}

@inproceedings{wang-etal-2024-scimon,
    title = "{S}ci{MON}: Scientific Inspiration Machines Optimized for Novelty",
    author = "Wang, Qingyun  and
      Downey, Doug  and
      Ji, Heng  and
      Hope, Tom",
    editor = "Ku, Lun-Wei  and
      Martins, Andre  and
      Srikumar, Vivek",
    booktitle = "Proceedings of the 62nd Annual Meeting of the Association for Computational Linguistics (Volume 1: Long Papers)",
    month = aug,
    year = "2024",
    address = "Bangkok, Thailand",
    publisher = "Association for Computational Linguistics",
    url = "https://aclanthology.org/2024.acl-long.18/",
    doi = "10.18653/v1/2024.acl-long.18",
    pages = "279--299"
}

@inproceedings{garikaparthi-etal-2025-iris,
    title = "{IRIS}: Interactive Research Ideation System for Accelerating Scientific Discovery",
    author = "Garikaparthi, Aniketh  and
      Patwardhan, Manasi  and
      Vig, Lovekesh  and
      Cohan, Arman",
    editor = "Mishra, Pushkar  and
      Muresan, Smaranda  and
      Yu, Tao",
    booktitle = "Proceedings of the 63rd Annual Meeting of the Association for Computational Linguistics (Volume 3: System Demonstrations)",
    month = jul,
    year = "2025",
    address = "Vienna, Austria",
    publisher = "Association for Computational Linguistics",
    url = "https://aclanthology.org/2025.acl-demo.57/",
    doi = "10.18653/v1/2025.acl-demo.57",
    pages = "592--603",
    ISBN = "979-8-89176-253-4"
}

@article{DBLP:journals/corr/abs-2402-10886,
  publtype={informal},
  author={Zhaolin Gao and Kianté Brantley and Thorsten Joachims},
  title={Reviewer2: Optimizing Review Generation Through Prompt Generation},
  year={2024},
  cdate={1704067200000},
  journal={CoRR},
  volume={abs/2402.10886},
  url={https://doi.org/10.48550/arXiv.2402.10886}
}

@misc{zhu2025deepreviewimprovingllmbasedpaper,
      title={DeepReview: Improving LLM-based Paper Review with Human-like Deep Thinking Process}, 
      author={Minjun Zhu and Yixuan Weng and Linyi Yang and Yue Zhang},
      year={2025},
      eprint={2503.08569},
      archivePrefix={arXiv},
      primaryClass={cs.CL},
      url={https://arxiv.org/abs/2503.08569}, 
}

@inproceedings{schmidgall-etal-2025-agent,
    title = "Agent Laboratory: Using {LLM} Agents as Research Assistants",
    author = "Schmidgall, Samuel  and
      Su, Yusheng  and
      Wang, Ze  and
      Sun, Ximeng  and
      Wu, Jialian  and
      Yu, Xiaodong  and
      Liu, Jiang  and
      Moor, Michael  and
      Liu, Zicheng  and
      Barsoum, Emad",
    editor = "Christodoulopoulos, Christos  and
      Chakraborty, Tanmoy  and
      Rose, Carolyn  and
      Peng, Violet",
    booktitle = "Findings of the Association for Computational Linguistics: EMNLP 2025",
    month = nov,
    year = "2025",
    address = "Suzhou, China",
    publisher = "Association for Computational Linguistics",
    url = "https://aclanthology.org/2025.findings-emnlp.320/",
    doi = "10.18653/v1/2025.findings-emnlp.320",
    pages = "5977--6043",
    ISBN = "979-8-89176-335-7"
}

@inproceedings{10.5555/3692070.3692738,
author = {Guo, Siyuan and Deng, Cheng and Wen, Ying and Chen, Hechang and Chang, Yi and Wang, Jun},
title = {DS-agent: automated data science by empowering large language models with case-based reasoning},
year = {2024},
publisher = {JMLR.org},
booktitle = {Proceedings of the 41st International Conference on Machine Learning},
articleno = {668},
numpages = {36},
location = {Vienna, Austria},
series = {ICML'24}
}

@misc{yetiştiren2023evaluatingcodequalityaiassisted,
      title={Evaluating the Code Quality of AI-Assisted Code Generation Tools: An Empirical Study on GitHub Copilot, Amazon CodeWhisperer, and ChatGPT}, 
      author={Burak Yetiştiren and Işık Özsoy and Miray Ayerdem and Eray Tüzün},
      year={2023},
      eprint={2304.10778},
      archivePrefix={arXiv},
      primaryClass={cs.SE},
      url={https://arxiv.org/abs/2304.10778}, 
}

@article{doi:10.1142/S0218194022500358,
author = {Lin, Jialiang and Wang, Yingmin and Yu, Yao and Zhou, Yu and Chen, Yidong and Shi, Xiaodong},
title = {Automatic Analysis of Available Source Code of Top Artificial Intelligence Conference Papers},
journal = {International Journal of Software Engineering and Knowledge Engineering},
volume = {32},
number = {07},
pages = {947-970},
year = {2022},
doi = {10.1142/S0218194022500358},
URL = { https://doi.org/10.1142/S0218194022500358},
eprint = { https://doi.org/10.1142/S0218194022500358}
}

@article{resnik2017reproducibility,
  title={Reproducibility and research integrity},
  author={Resnik, David B and Shamoo, Adil E},
  journal={Accountability in research},
  volume={24},
  number={2},
  pages={116--123},
  year={2017},
  publisher={Taylor \& Francis}
}

@misc{kim2025reproductionreplicationevaluatingresearch,
      title={From Reproduction to Replication: Evaluating Research Agents with Progressive Code Masking}, 
      author={Gyeongwon James Kim and Alex Wilf and Louis-Philippe Morency and Daniel Fried},
      year={2025},
      eprint={2506.19724},
      archivePrefix={arXiv},
      primaryClass={cs.AI},
      url={https://arxiv.org/abs/2506.19724}, 
}

@misc{wei2025browsecompsimplechallengingbenchmark,
      title={BrowseComp: A Simple Yet Challenging Benchmark for Browsing Agents}, 
      author={Jason Wei and Zhiqing Sun and Spencer Papay and Scott McKinney and Jeffrey Han and Isa Fulford and Hyung Won Chung and Alex Tachard Passos and William Fedus and Amelia Glaese},
      year={2025},
      eprint={2504.12516},
      archivePrefix={arXiv},
      primaryClass={cs.CL},
      url={https://arxiv.org/abs/2504.12516}, 
}

@article{dragomir2025decentralized,
  title={A Decentralized Hierarchical Multi-Agent Framework for Smart Grid Sustainable Energy Management},
  author={Dragomir, Otilia Elena and Dragomir, Florin},
  journal={Sustainability},
  volume={17},
  number={12},
  pages={5423},
  year={2025},
  publisher={MDPI}
}

@misc{feng2024hierarchicalconsensusbasedmultiagentreinforcement,
      title={Hierarchical Consensus-Based Multi-Agent Reinforcement Learning for Multi-Robot Cooperation Tasks}, 
      author={Pu Feng and Junkang Liang and Size Wang and Xin Yu and Xin Ji and Yiting Chen and Kui Zhang and Rongye Shi and Wenjun Wu},
      year={2024},
      eprint={2407.08164},
      archivePrefix={arXiv},
      primaryClass={cs.AI},
      url={https://arxiv.org/abs/2407.08164}, 
}

@misc{lallement2014hatphtnplannerrobotics,
      title={HATP: An HTN Planner for Robotics}, 
      author={Raphaël Lallement and Lavindra de Silva and Rachid Alami},
      year={2014},
      eprint={1405.5345},
      archivePrefix={arXiv},
      primaryClass={cs.RO},
      url={https://arxiv.org/abs/1405.5345}, 
}

@article{10.5555/3546258.3546422,
author = {Pineau, Joelle and Vincent-Lamarre, Philippe and Sinha, Koustuv and Larivi\`{e}re, Vincent and Beygelzimer, Alina and d'Alch\'{e}-Buc, Florence and Fox, Emily and Larochelle, Hugo},
title = {Improving reproducibility in machine learning research (a report from the NeurIPS 2019 reproducibility program)},
year = {2021},
issue_date = {January 2021},
publisher = {JMLR.org},
volume = {22},
number = {1},
issn = {1532-4435},
journal = {J. Mach. Learn. Res.},
month = jan,
articleno = {164},
numpages = {20}
}

@misc{wang2020romamultiagentreinforcementlearning,
      title={ROMA: Multi-Agent Reinforcement Learning with Emergent Roles}, 
      author={Tonghan Wang and Heng Dong and Victor Lesser and Chongjie Zhang},
      year={2020},
      eprint={2003.08039},
      archivePrefix={arXiv},
      primaryClass={cs.MA},
      url={https://arxiv.org/abs/2003.08039}, 
}

@inproceedings{yuan-etal-2025-dolphin,
    title = "Dolphin: Moving Towards Closed-loop Auto-research through Thinking, Practice, and Feedback",
    author = "Yuan, Jiakang  and
      Yan, Xiangchao  and
      Zhang, Bo  and
      Chen, Tao  and
      Shi, Botian  and
      Ouyang, Wanli  and
      Qiao, Yu  and
      Bai, Lei  and
      Zhou, Bowen",
    editor = "Che, Wanxiang  and
      Nabende, Joyce  and
      Shutova, Ekaterina  and
      Pilehvar, Mohammad Taher",
    booktitle = "Proceedings of the 63rd Annual Meeting of the Association for Computational Linguistics (Volume 1: Long Papers)",
    month = jul,
    year = "2025",
    address = "Vienna, Austria",
    publisher = "Association for Computational Linguistics",
    url = "https://aclanthology.org/2025.acl-long.1056/",
    doi = "10.18653/v1/2025.acl-long.1056",
    pages = "21768--21789",
    ISBN = "979-8-89176-251-0"
}

@inproceedings{liang-etal-2024-encouraging,
    title = "Encouraging Divergent Thinking in Large Language Models through Multi-Agent Debate",
    author = "Liang, Tian  and
      He, Zhiwei  and
      Jiao, Wenxiang  and
      Wang, Xing  and
      Wang, Yan  and
      Wang, Rui  and
      Yang, Yujiu  and
      Shi, Shuming  and
      Tu, Zhaopeng",
    editor = "Al-Onaizan, Yaser  and
      Bansal, Mohit  and
      Chen, Yun-Nung",
    booktitle = "Proceedings of the 2024 Conference on Empirical Methods in Natural Language Processing",
    month = nov,
    year = "2024",
    address = "Miami, Florida, USA",
    publisher = "Association for Computational Linguistics",
    url = "https://aclanthology.org/2024.emnlp-main.992/",
    doi = "10.18653/v1/2024.emnlp-main.992",
    pages = "17889--17904"
}

@inproceedings{10.5555/3692070.3692537,
author = {Du, Yilun and Li, Shuang and Torralba, Antonio and Tenenbaum, Joshua B. and Mordatch, Igor},
title = {Improving factuality and reasoning in language models through multiagent debate},
year = {2024},
publisher = {JMLR.org},
booktitle = {Proceedings of the 41st International Conference on Machine Learning},
articleno = {467},
numpages = {31},
location = {Vienna, Austria},
series = {ICML'24}
}

@inproceedings{10.5555/3692070.3692275,
author = {Cai, Diana and Modi, Chirag and Pillaud-Vivien, Loucas and Margossian, Charles C. and Gower, Robert M. and Blei, David M. and Saul, Lawrence K.},
title = {Batch and match: black-box variational inference with a score-based divergence},
year = {2024},
publisher = {JMLR.org},
booktitle = {Proceedings of the 41st International Conference on Machine Learning},
articleno = {205},
numpages = {40},
location = {Vienna, Austria},
series = {ICML'24}
}

@misc{
sanchez2024stay,
title={Stay on Topic with Classifier-Free Guidance},
author={Guillaume Sanchez and Alexander Spangher and Honglu Fan and Elad Levi and Pawan Sasanka Ammanamanchi and Stella Biderman},
year={2024},
url={https://openreview.net/forum?id=RmRA7Q0lwQ}
}

@misc{hong2025onesizefitsallinversionlearninghighly,
      title={Beyond One-Size-Fits-All: Inversion Learning for Highly Effective NLG Evaluation Prompts}, 
      author={Hanhua Hong and Chenghao Xiao and Yang Wang and Yiqi Liu and Wenge Rong and Chenghua Lin},
      year={2025},
      eprint={2504.21117},
      archivePrefix={arXiv},
      primaryClass={cs.CL},
      url={https://arxiv.org/abs/2504.21117}, 
}

@misc{wu2026largescaleterminalagentictrajectory,
      title={Large-Scale Terminal Agentic Trajectory Generation from Dockerized Environments}, 
      author={Siwei Wu and Yizhi Li and Yuyang Song and Wei Zhang and Yang Wang and Riza Batista-Navarro and Xian Yang and Mingjie Tang and Bryan Dai and Jian Yang and Chenghua Lin},
      year={2026},
      eprint={2602.01244},
      archivePrefix={arXiv},
      primaryClass={cs.CL},
      url={https://arxiv.org/abs/2602.01244}, 
}
% \label{sec:appendix}

% This is an appendix.

\appendix

\onecolumn
\section{Linear Regression Plots}
\label{app:linreg}
\vspace{10em}
\begin{figure}[htb]
    \centering
    \begin{subfigure}[t]{0.48\textwidth}
        \centering 
        \includegraphics[width=0.9\textwidth]{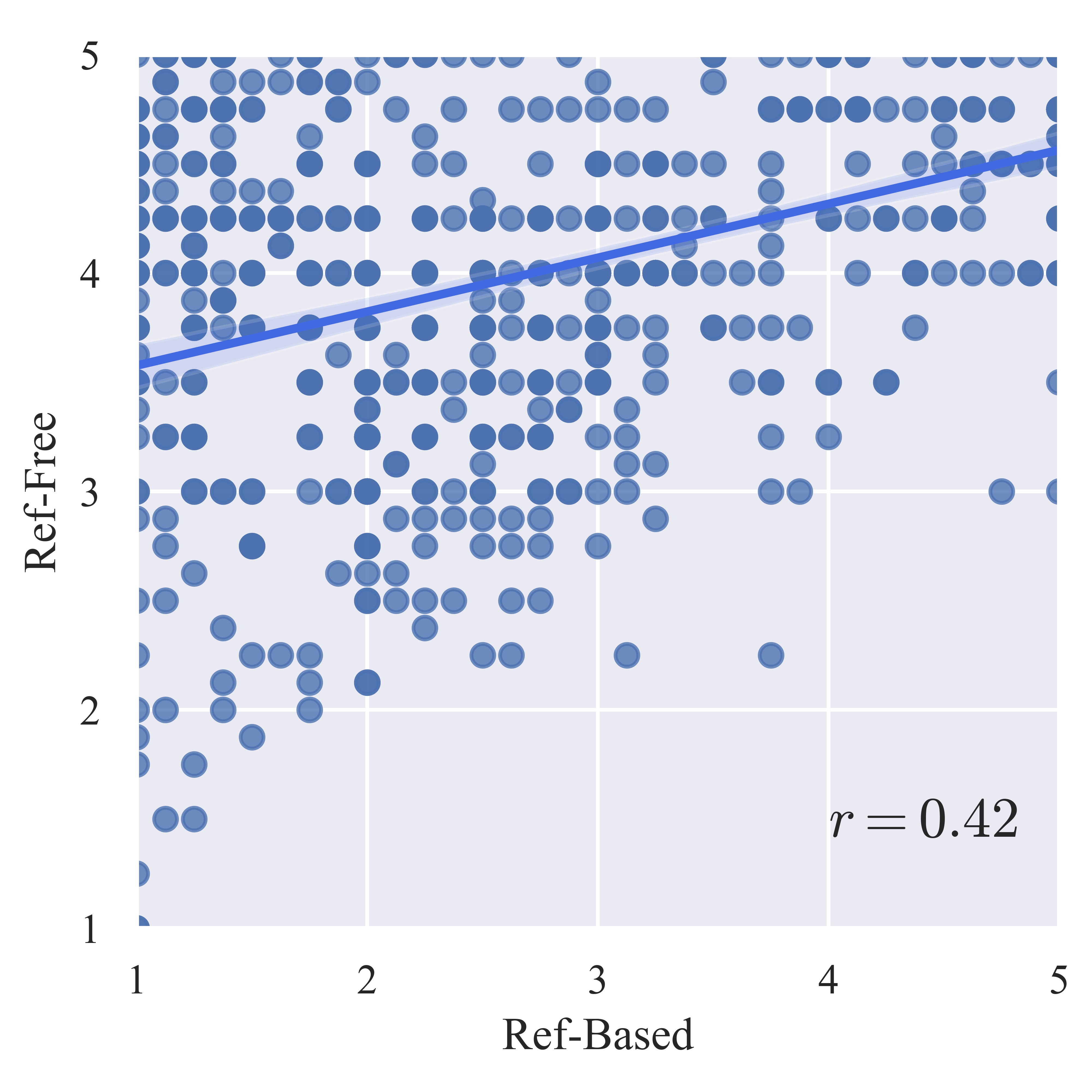}
        \caption{Ref-Based vs. Ref-Free}
        \label{fig:ref_free}
    \end{subfigure}
    \begin{subfigure}[t]{0.48\textwidth}
        \centering 
        \includegraphics[width=0.9\textwidth]{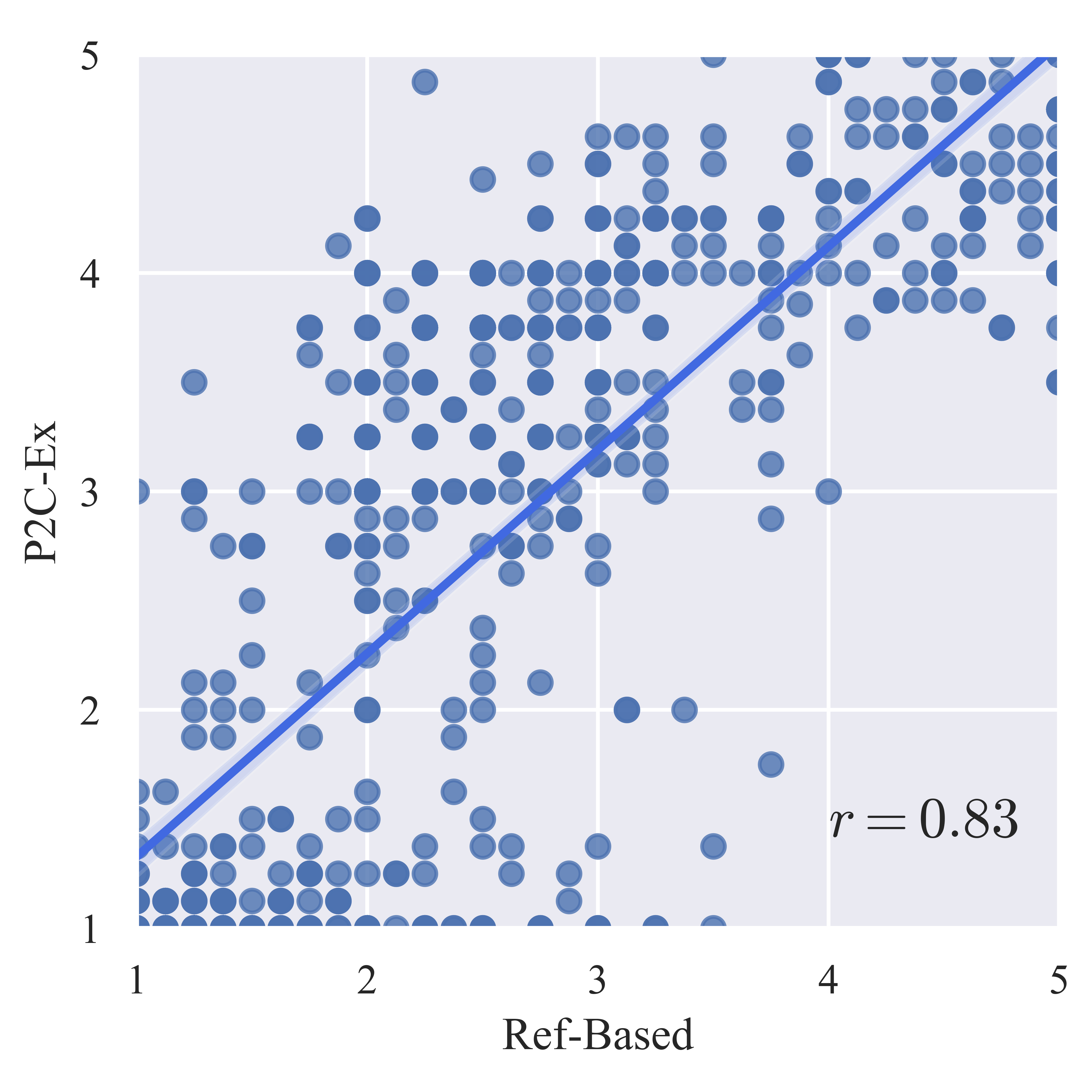}
        \caption{Ref-Based vs. P2C-Ex}
        \label{fig:p2c_ex}
    \end{subfigure}
    \caption{Repository-level linear regression plots. Points represent the scoring of repositories, with the x-axis showing the reference-based evaluation score and the y-axis showing the corresponding reference-free score.}
    \label{fig:regplot}
\end{figure}
\clearpage
\twocolumn
%\section{Human Evaluation}
%\section{Error Analysis}
\clearpage
\section{Reproduction Case Study Details}

We present representative reproduction outcomes produced by PaperCoder and \ourwork for the ICML 2024 paper \textit{Batch and Match: Black-Box Variational Inference with a Score-Based Divergence}~\cite{10.5555/3692070.3692275}, which is included in the PaperBench~\cite{starace2025paperbenchevaluatingaisability} benchmark. The paper introduces Batch and Match (BaM), a black-box variational inference method based on a score-based divergence, and evaluates its performance on both Gaussian and non-Gaussian distributions arising from posterior inference in hierarchical models as well as deep generative models.

The objective of experiment reproduction frameworks is to understand the original methodology and experimental design, generate the corresponding code, and ultimately reproduce the reported results. Figure~\ref{fig:repo_examples} compares the code repositories produced by PaperCoder and \ourwork. Figure~\ref{fig:manager_prompt} illustrates a case in which the planning manager agent $\mathcal{A}\mathrm{plan}$ re-invokes the overall planning agent $\mathcal{A}\mathrm{overall}$ due to insufficient implementation detail. Finally, Figure~\ref{fig:plan_examples} contrasts the implementation roadmaps generated by the two frameworks.

\label{app:case_examples}
\begin{figure*}[b]

\tcbset{
    colback=gray!5!white,
%    width=0.95\columnwidth,
    fontupper=\small,
    left=1pt,
    right=1pt,
    valign=center,
    before=\vspace{0pt},
    after=\vspace{0pt}
}
    \centering
    \begin{subfigure}[c]{0.48\textwidth}
       \begin{tcolorbox}[]
       - bam\_algorithm.py\\
       - config.yaml\\
       - evaluation\_metrics.py\\
       - experiments.py\\
       - main.py\\
       - target\_distributions.py\\
       - utils.py\\
       - variational\_family.py\\
        \end{tcolorbox}
        \caption{PaperCoder}
        \label{fig:ref_free}
    \end{subfigure}
    \begin{subfigure}[c]{0.48\textwidth}
        \centering 
           \begin{tcolorbox}[]
            - algorithms/\\
            \tcbtab- bam.py\\
            \tcbtab- baselines.py\\
            - experiments/\\
            \tcbtab- gaussian\_experiments.py\\
            - metrics/\\
            \tcbtab- divergences.py\\
            \tcbtab- errors.py\\
            - targets/\\
            \tcbtab- gaussian.py\\
            \tcbtab- non\_gaussian.py\\
            - utils/\\
            \tcbtab- config.py\\
            \tcbtab- experiment\_tracker.py\\
            \tcbtab- matrix\_ops.py\\
            - basic\_gaussian\_exp.py\\ 
            - main.py\\
            - minimal\_exp.py\\
            - README.md \\
            - requirements.txt \\
            - simple\_test.py\\
            - test\_bam.py\\
            \end{tcolorbox}
        \caption{\ourwork}
        \label{fig:p2c_ex}
    \end{subfigure}
    \vspace{-1em}
    \caption{Structure illustrations of code repositories generated by PaperCoder and \ourwork using \hyphentt{DeepSeek-v3.1-Terminus} for the paper~\cite{10.5555/3692070.3692275}.}
    \label{fig:repo_examples}
    \vspace{-1em}
\end{figure*}
\clearpage
\onecolumn

\begin{figure}[]
\centering
\tcbset{
    colback=gray!5!white,
    width=0.95\columnwidth,
    fontupper=\small,
    left=1pt,
    right=1pt,
    valign=center,
    before=\vspace{0pt},
    after=\vspace{0pt}
}
\begin{tcolorbox}[]
The file you generated is not sufficiently detailed. Please regenerate it with comprehensive, detailed content that properly addresses the requirements.

The plan file should contain:

1. Detailed step-by-step implementation plan for each experiment section:

\tcbtab- Gaussian targets (Section 5.1)

\tcbtab- Non-Gaussian targets (Section 5.1)

\tcbtab- PosteriorDB Bayesian models (Section 5.2)

\tcbtab- Deep generative models (VAE application, Section 5.3)

2. Specific milestones with dates and deliverables:

\tcbtab- Step-by-step breakdown of implementation tasks

\tcbtab- Clear deliverables for each milestone

\tcbtab- Integration and testing phases

3. Resource allocation (computational, time, personnel):

\tcbtab- Computational resources required (GPU, CPU, memory)

\tcbtab- Time estimates for each phase

\tcbtab- Personnel roles and responsibilities

4. Risk assessment and mitigation strategies:

\tcbtab- Technical risks (algorithm convergence, implementation complexity)

\tcbtab- Resource risks (computational requirements, time constraints)

\tcbtab- Mitigation strategies for each identified risk

The plan should be comprehensive and ready for implementation. The current content is insufficient.

\end{tcolorbox}
\caption{The prompt from the planning manager agent $\mathcal{A}_\mathrm{plan}$ to re-invoke the subordinate agent $\mathcal{A}_\mathrm{overall}$}
\label{fig:manager_prompt}
\end{figure}

\begin{figure}[htb]

\tcbset{
    colback=gray!5!white,
%    width=0.95\columnwidth,
    fontupper=\small,
    left=1pt,
    right=1pt,
    valign=center,
    before=\vspace{0pt},
    after=\vspace{0pt}
}
    \centering
    \begin{subfigure}[c]{0.48\textwidth}
       \begin{tcolorbox}[]
\#\# Implementation Roadmap

\#\#\# Core Modules

1. **BaM algorithm class**

2. **Score function interface** for target distributions

3. **Gaussian variational family** with mean/covariance parameters

4. **Matrix operations** (quadratic equation solver, low-rank optimizations)

\#\#\# Experiment Modules

1. **Synthetic target generators** (Gaussian, sinh-arcsinh)

2. **Benchmark models** (ADVI, GSM implementations)

3. **Evaluation metric calculators**

4. **Visualization tools** for results

\#\#\# Testing Strategy

1. **Unit tests** for core algorithm components

2. **Gradient checking** for score computations

3. **Convergence verification** on simple Gaussian targets

4. **Reproduction of paper figures** with statistical significance
        \end{tcolorbox}
        \caption{PaperCoder}
        \label{fig:ref_free}
    \end{subfigure}
    \begin{subfigure}[c]{0.48\textwidth}
        \centering 
           \begin{tcolorbox}[]
\#\# Specific Milestones and Deliverables

\#\#\# Phase 1: Core Implementation 

Algorithm Core

- [ ] Implement BaM algorithm with full covariance

- [ ] Implement low-rank solver for B < D

- [ ] Create Gaussian target test suite

Baseline Implementations

- [ ] Implement ADVI, GSM, Score/Fisher methods

- [ ] Create unified experiment framework

**Deliverable**: Working implementation of all algorithms with basic testing

\#\#\# Phase 2: Gaussian \& Non-Gaussian Experiments 

Gaussian Targets

- [ ] Run experiments for D = 4, 16, 64, 256

- [ ] Generate convergence plots (forward/reverse KL)

- [ ] Statistical analysis (10 runs per configuration)

Non-Gaussian Targets

- [ ] Implement sinh-arcsinh distribution

- [ ] Run experiments with varying skew/tail parameters

- [ ] Compare convergence across methods

**Deliverable**: Complete Section 5.1 reproduction with statistical analysis

\#\#\# Phase 3: Bayesian Models

BridgeStan Integration

- [ ] Integrate with PosteriorDB models

- [ ] Generate HMC reference samples

- [ ] Implement relative error metrics

PosteriorDB Experiments

- [ ] Run experiments on eight\_schools, garch, gp\_pois\_regr

- [ ] Compare relative mean/SD errors

- [ ] Statistical analysis (5 runs per model)

Complete Section 5.2 reproduction with error analysis

\#\#\# Phase 4: Deep Generative Models

VAE Implementation

- [ ] Implement 5-layer convolutional encoder/decoder

- [ ] Pre-train VAE on image dataset (100 epochs)

- [ ] Validate reconstruction quality

Posterior Inference Experiments

- [ ] Run BaM on test image posteriors

- [ ] Compare reconstruction MSE across methods

- [ ] Analyze convergence behavior

**Deliverable**: Complete Section 5.3 reproduction with VAE application

\#\#\# Phase 5: Integration \& Final Analysis 

Integration Testing

- [ ] Cross-validate all implementations

- [ ] Performance optimization

- [ ] Code documentation

Final Analysis \& Reporting

- [ ] Statistical summary of all experiments

- [ ] Comparison with paper results

- [ ] Final report preparation

**Deliverable**: Comprehensive reproduction report
            \end{tcolorbox}
        \caption{\ourwork}
        \label{fig:p2c_ex}
    \end{subfigure}
    \caption{The implementation roadmap section in the plans generated by PaperCoder and \ourwork using \hyphentt{DeepSeek-v3.1-Terminus} for the paper~\cite{10.5555/3692070.3692275}.}
    \label{fig:plan_examples}
\end{figure}

\twocolumn
\clearpage
\section{Failure Mode Example}
\label{app:failure_analysis}
Figure~\ref{fig:repo_examples} illustrates the structure of a code repository generated by \ourwork with \hyphentt{DeepSeek-v3.1-Terminus} as the backbone for reproducing the paper \textit{Stay on Topic with Classifier-Free Guidance}~\cite{sanchez2024stay} from PaperBench. In this case, \ourwork produces a well-structured and complete codebase, attaining a high score of 69.7\% on the PaperBench-CodeDev subset, which evaluates only the Code Development requirements. However, on the full PaperBench benchmark, which additionally assesses Execution and Result matching, the score drops substantially to 11.3\%. Closer manual inspection reveals that the execution agent struggles to correctly run the main program due to improper inter-file imports across components (e.g., \texttt{model\_manager.py} and \texttt{cfg\_engine.py}). This discrepancy highlights both the proficiency of current models and frameworks in code development and their failure in execution, stemming from weak cross-file coding consistency and inadequate coordination among interdependent modules.

\begin{figure}[h]
\tcbset{
    colback=gray!5!white,
    width=0.8\columnwidth,
%    fontupper=\small,
    left=5pt,
    right=5pt,
    valign=center,
    before=\vspace{0pt},
    after=\vspace{0pt}
}
    \centering
\begin{tcolorbox}[]
code/\\
\tcbtab cfg\_engine.py\\
\tcbtab code\_generation\_evaluator.py\\
\tcbtab config.py\\
\tcbtab data\_loader.py\\
\tcbtab experiment\_runner.py\\
\tcbtab flops\_analyzer.py\\
\tcbtab main.py\\
\tcbtab model\_manager.py\\
\tcbtab results\_analyzer.py\\
\tcbtab utils.py\\
\tcbtab zero\_shot\_evaluator.py\\
        \end{tcolorbox}
    \caption{Structure illustration of the code repository generated by \ourwork using \hyphentt{DeepSeek-v3.1-Terminus} for the paper~\cite{sanchez2024stay}.}
    \label{fig:repo_examples}
\end{figure}

\onecolumn
\clearpage
\section{Paper2Code Evaluation Prompts}
\label{app:p2c_prompts}
\tcbset{
    colback=gray!5!white,
    fontupper=\scriptsize,
    left=1pt,
    right=1pt,
    before=\vspace{0pt},
    after=\vspace{0pt}
}

\begin{figure}[h]
    \centering
    \begin{tcolorbox}
You will be given a research paper along with two corresponding code repositories: a gold repository and a target repository.

Your task is to compare the target repository against the gold repository, rate the target repository on one metric, and provide a critique highlighting key differences.

Please make sure you read and understand these instructions carefully. Keep this document open while reviewing, and refer to it as needed.

---

Evaluation Criteria:

Correctness (1-5): The quality of the target repository in accurately implementing the paper’s concepts, methodology, and algorithms without logical errors, as compared to the gold repository. Additionally, provide a critique focusing on the completeness, accuracy, and implementation choices made in the target repository relative to the gold repository.

1: Very Poor. The target repository does not correctly implement the core concepts, methodology, or algorithms from the paper. Major logical errors or missing components are present, especially when compared to the gold repository.
2: Poor. The target repository attempts to implement the paper’s concepts but contains significant mistakes or missing components, making the implementation incorrect when compared to the gold repository.
3: Fair. Some core components and concepts are correctly implemented in the target repository, but there are notable logical errors or inaccuracies compared to the gold repository.
4: Good. The target repository correctly implements the key components and methodology, with only minor inaccuracies or deviations from the gold repository.
5: Excellent. The target repository fully and accurately implements all relevant key components, methodology, and algorithms from the paper, matching the quality of the gold repository.

---

Evaluation Steps  

1. Identify Key Aspects of the Paper: Carefully read the research paper to understand its core concepts, methodology, and algorithms. Pay close attention to the key aspects that are crucial for implementing the paper’s results (e.g., specific algorithms, data preprocessing steps, evaluation protocols).

2. Analyze the Gold Repository: Examine the gold repository to understand how these key aspects have been implemented. Use the gold repository as a reference for how the paper’s methodology should be translated into code. Note the completeness, accuracy, and design choices in the gold repository that faithfully represent the paper’s concepts.

3. Examine the Target Repository: Analyze the target repository to assess how well it implements the key aspects of the paper. Reference the gold repository as a guide for understanding these key aspects in the target repository. Focus on whether the target repository’s core logic, algorithms, and structure align with the methodology and experiments described in the paper.

4. Identify Logical Errors and Deviations: Check for logical errors, missing steps, or deviations from the paper’s methodology. Note any incorrect representations, inconsistencies, or incomplete implementations that could affect the correctness of the target repository.

5. Provide a Critique: Consider both the completeness and accuracy of the implementation relative to the paper’s goals and the gold repository’s standard. You do not need to analyze minor details like logging functions, script organization, or documentation quality. Instead, concentrate on the correctness of the logic and implementation that ensures the core concepts from the paper are fully reflected in the target repository. For each mismatch or deviation in implementation, note down specific critiques comparing relevant functions in the target repository to the corresponding functions in the gold repository. Highlight incorrect logic, missing steps, or deviations that affect the correct implementation of the paper’s methodology.

6. Assess the Correctness: Determine whether the target repository includes all the critical elements described in the paper and implemented in the gold repository. Identify missing components, significant deviations, or incorrect implementations that could affect the correctness of the target repository.

7. Assign a Score: Based on your evaluation, provide a critique and assign a correctness score from 1 to 5 for the target repository, reflecting how well it implements the key aspects of the paper refer to the gold repository. Include a detailed critique in the specified JSON format.

---

Severity Level:  

Each identified critique will be assigned a severity level based on its impact on the correctness of the methodology implementation.  

- High: Missing or incorrect implementation of the paper’s core concepts, major loss functions, or experiment components that are fundamental to reproducing the paper’s methodology.  

\tcbtab- Example: The main algorithm is missing or fundamentally incorrect.  

- Medium: Issues affecting training logic, data preprocessing, or other core functionalities that significantly impact performance but do not completely break the system.  

\tcbtab- Example: Improper training loop structure, incorrect data augmentation, or missing essential components in data processing.  

- Low: Errors in specific features that cause deviations from expected results but can be worked around with modifications. Any errors in the evaluation process belong to this category unless they impact the core concepts. These include minor issues like logging, error handling mechanisms, configuration settings, evaluation steps that do not alter the fundamental implementation and additional implementations not explicitly stated in the paper.

\tcbtab- Example: Suboptimal hyperparameter initialization, incorrect learning rate schedule, inaccuracies in evaluation metrics, using a different random seed, variations in batch processing, different weight initialization, issues in result logging or reporting, variations in evaluation dataset splits, improper error handling in non-critical steps, mismatches in secondary evaluation criteria, or additional implementation details not specified in the paper that do not interfere with core results.

---

Example JSON format:

```json\\
\{\\
\tcbtab"critique\_list": [\\
\tcbtab\tcbtab\{\\
\tcbtab\tcbtab"gold\_file\_name":\\ \tcbtab\tcbtab\tcbtab"preprocessing.py",\\
\tcbtab\tcbtab\tcbtab"gold\_func\_name": "data\_process",\\
\tcbtab\tcbtab\tcbtab"target\_file\_name": "dataset.py",\\
\tcbtab\tcbtab\tcbtab"target\_func\_name": "train\_preprocess",\\
\tcbtab\tcbtab\tcbtab"severity\_level": "medium",\\
\tcbtab\tcbtab\tcbtab"critique": "A critique of the target repository's file with reference to the gold repository."\\
\tcbtab\tcbtab\}\\
\tcbtab],\\
\tcbtab"score": 2\\
\}\\
```

---

Sample:

Research Paper:

\{\{Paper\}\}

Gold Repository:

\{\{GoldCode\}\}

Target Repository:

\{\{Code\}\}

---

Please provide critique of the target repository and a single numerical rating (1, 2, 3, 4, or 5) based on the quality of the sample, following the Example JSON format, without any additional commentary, formatting, or chattiness.
    \end{tcolorbox}
    \vspace{-2em}
    \caption{The original prompt for reference-based evaluation}
    \label{fig:ref-based_prompt}
    \vspace{-5cm}
\end{figure}

\clearpage
\begin{figure}[h]
    \centering
    \begin{tcolorbox}
You will be given a research paper along with its corresponding code repository.

Your task is to rate the code repository on one metric and provide a critique highlighting key differences.

Please make sure you read and understand these instructions carefully. Keep this document open while reviewing, and refer to it as needed.

---

Evaluation Criteria:

Correctness (1-5): The quality of the repository in accurately implementing the paper’s concepts, methodology, and algorithms without logical errors. Additionally, provide a critique focusing on the completeness, accuracy, and implementation choices made in the repository relative to the methodology and algorithms described in the paper.

1: Very Poor. The repository does not correctly implement the core concepts, methodology, or algorithms from the paper. Major logical errors or missing components are present.

2: Poor. The repository attempts to implement the paper’s concepts but contains significant mistakes or missing components, making the implementation incorrect.

3: Fair. Some core components and concepts are correctly implemented, but there are notable logical errors or inaccuracies in the methodology.

4: Good. The repository correctly implements the key components and methodology, with only minor inaccuracies that do not significantly affect correctness.

5: Excellent. The repository fully and accurately implements all key components, methodology, and algorithms from the paper without logical errors.

---

Evaluation Steps  

1. Identify Key Aspects of the Paper: Carefully read the paper to understand its core concepts, methodology, and algorithms. Pay close attention to key aspects crucial for implementing the paper’s results (e.g., specific algorithms, data preprocessing steps, evaluation protocols).

2. Examine the Code Repository: Analyze the repository to determine how well it implements the key aspects of the paper. Focus on whether the repository’s core logic, algorithms, and structure align with the methodology and experiments described in the paper.

3. Identify Logical Errors and Deviations: Check for logical errors, missing steps, or deviations from the paper’s methodology. Note any incorrect representations, inconsistencies, or incomplete implementations that could affect the correctness of the repository.

4. Provide a Critique: Consider the completeness and accuracy of the implementation relative to the paper’s goals. You do not need to analyze minor details like logging functions, script organization, or documentation quality. Instead, concentrate on the correctness of the logic and implementation to ensure the core concepts from the paper are fully reflected in the repository. For each identified issue, write a detailed critique specifying the affected files and functions in the repository. Highlight missing or incorrectly implemented steps that impact the correctness and alignment with the paper’s methodology.

5. Assess Completeness and Accuracy: Evaluate the repository for its completeness and accuracy relative to the paper’s methodology. Ensure that all critical components—such as data preprocessing, core algorithms, and evaluation steps—are implemented and consistent with the paper’s descriptions.

6. Assign a Score: Based on your evaluation, provide a critique and assign a correctness score from 1 to 5 for the repository, reflecting how well it implements the key aspects of the paper. Include a detailed critique in the specified JSON format.

---

Severity Level:  

Each identified critique will be assigned a severity level based on its impact on the correctness of the methodology implementation.  

- High: Missing or incorrect implementation of the paper’s core concepts, major loss functions, or experiment components that are fundamental to reproducing the paper’s methodology.  

\tcbtab- Example: The main algorithm is missing or fundamentally incorrect.  
  
- Medium: Issues affecting training logic, data preprocessing, or other core functionalities that significantly impact performance but do not completely break the system.  

\tcbtab- Example: Improper training loop structure, incorrect data augmentation, or missing essential components in data processing.  
  
- Low: Errors in specific features that cause deviations from expected results but can be worked around with modifications. Any errors in the evaluation process belong to this category unless they impact the core concepts. These include minor issues like logging, error handling mechanisms, configuration settings, evaluation steps that do not alter the fundamental implementation and additional implementations not explicitly stated in the paper.

\tcbtab- Example: Suboptimal hyperparameter initialization, incorrect learning rate schedule, inaccuracies in evaluation metrics, using a different random seed, variations in batch processing, different weight initialization, issues in result logging or reporting, variations in evaluation dataset splits, improper error handling in non-critical steps, mismatches in secondary evaluation criteria, or additional implementation details not specified in the paper that do not interfere with core results.

---

Example JSON format:  
```json
\{\\
\tcbtab"critique\_list": [\\
\tcbtab\tcbtab\{\\
\tcbtab\tcbtab\tcbtab"file\_name": "dataset.py",\\
\tcbtab\tcbtab\tcbtab"func\_name": "train\_preprocess",\\
\tcbtab\tcbtab\tcbtab"severity\_level": "medium",\\
\tcbtab\tcbtab\tcbtab"critique": "A critique of the target repository's file."\\
\tcbtab\tcbtab\},\\
\tcbtab\tcbtab\{\\
\tcbtab\tcbtab\tcbtab"file\_name": "metrics.py",\\
\tcbtab\tcbtab\tcbtab"func\_name": "f1\_at\_k",\\
\tcbtab\tcbtab\tcbtab"severity\_level": "low",\\
\tcbtab\tcbtab\tcbtab"critique": "A critique of the target repository's file."\\
\tcbtab\tcbtab\}\\
\tcbtab],\\
\tcbtab"score": 2\\
\}\\
```

---

Sample:

Research Paper:

\{\{Paper\}\}

Code Repository:

\{\{Code\}\}

---

Please provide a critique list for the code repository and a single numerical rating (1, 2, 3, 4, or 5) based on the quality of the sample, following the Example JSON format, without any additional commentary, formatting, or chattiness.
    \end{tcolorbox}
    \vspace{-2em}
    \caption{The original prompt for reference-free evaluation}
    \label{fig:ref-free_prompt}
    \vspace{-5cm}
\end{figure}

\begin{figure}[h]
    \centering
    \begin{tcolorbox}
You will be given a research paper along with its corresponding code repository.

Your task is to rate the code repository on one metric and provide a critique highlighting key differences.

Please make sure you read and understand these instructions carefully. Keep this document open while reviewing, and refer to it as needed.

---

Evaluation Criteria:

Correctness (1-5): The quality of the repository in accurately implementing the paper’s concepts, methodology, and algorithms without logical errors. Additionally, provide a critique focusing on the completeness, accuracy, and implementation choices made in the repository relative to the methodology and algorithms described in the paper.

1: Very Poor. The repository does not correctly implement the core concepts, methodology, or algorithms from the paper. Major logical errors or missing components are present.

2: Poor. The repository attempts to implement the paper’s concepts but contains significant mistakes or missing components, making the implementation incorrect.

3: Fair. Some core components and concepts are correctly implemented, but there are notable logical errors or inaccuracies in the methodology.

4: Good. The repository correctly implements the key components and methodology, with only minor inaccuracies that do not significantly affect correctness.

5: Excellent. The repository fully and accurately implements all key components, methodology, and algorithms from the paper without logical errors.

---

Evaluation Steps  

1. Identify Key Aspects of the Paper: Carefully read the paper to understand its core concepts, methodology, and algorithms. Pay close attention to key aspects crucial for implementing the paper’s results (e.g., specific algorithms, data preprocessing steps, evaluation protocols).

2. Examine the Code Repository: Analyze the repository to determine how well it implements the key aspects of the paper. Focus on whether the repository’s core logic, algorithms, and structure align with the methodology and experiments described in the paper.

3. Identify Logical Errors and Deviations: Check for logical errors, missing steps, or deviations from the paper’s methodology. Note any incorrect representations, inconsistencies, or incomplete implementations that could affect the correctness of the repository.

4. Provide a Critique: Consider the completeness and accuracy of the implementation relative to the paper’s goals. You do not need to analyze minor details like logging functions, script organization, or documentation quality. Instead, concentrate on the correctness of the logic and implementation to ensure the core concepts from the paper are fully reflected in the repository. For each identified issue, write a detailed critique specifying the affected files and functions in the repository. 

5. Assess Completeness and Accuracy: Evaluate the repository for its completeness and accuracy relative to the paper’s methodology. Ensure that all critical components—such as data preprocessing, core algorithms, and evaluation steps—are implemented and consistent with the paper’s descriptions.

\darkred{6. Code verification: Verify that all key components expected from the paper are fully implemented with codes in the repository. Evaluate completeness holistically rather than limiting your review to existing files. In your critique, explicitly identify any missing components, absent implementations, or deviations from expected behavior, including cases where functionality is described only in documentation (e.g., README, config files) but not implemented. Any such gaps that impact methodological correctness or alignment with the paper should be called out and assigned a high-severity critique.}

7. Assign a Score: Based on your evaluation, provide a critique and assign a correctness score from 1 to 5 for the repository, reflecting how well it implements the key aspects of the paper. Include a detailed critique in the specified JSON format.

---

Severity Level:  

Each identified critique will be assigned a severity level based on its impact on the correctness of the methodology implementation.  

- High: Missing or incorrect implementation of the paper’s core concepts, major loss functions, or experiment components that are fundamental to reproducing the paper’s methodology.  
  - Example: The main algorithm is missing, not implemented, implemented only as a descriptive note or plan without executable code, or fundamentally incorrect.  
- Medium: Issues affecting training logic, data preprocessing, or other core functionalities that significantly impact performance but do not completely break the system.  
  - Example: Improper training loop structure, incorrect data augmentation, or missing essential components in data processing.  
- Low: Errors in specific features that cause deviations from expected results but can be worked around with modifications. Any errors in the evaluation process belong to this category unless they impact the core concepts. These include minor issues like logging, error handling mechanisms, configuration settings, evaluation steps that do not alter the fundamental implementation and additional implementations not explicitly stated in the paper.
  - Example: Suboptimal hyperparameter initialization, incorrect learning rate schedule, inaccuracies in evaluation metrics, using a different random seed, variations in batch processing, different weight initialization, issues in result logging or reporting, variations in evaluation dataset splits, improper error handling in non-critical steps, mismatches in secondary evaluation criteria, or additional implementation details not specified in the paper that do not interfere with core results.

---

Example JSON format:  
```json\\
\{\\
\tcbtab"critique\_list": [\\
\tcbtab\tcbtab\{\\
\tcbtab\tcbtab\tcbtab"file\_name": "dataset.py",\\
\tcbtab\tcbtab\tcbtab"func\_name": "train\_preprocess",\\
\tcbtab\tcbtab\tcbtab"severity\_level": "medium",\\
\tcbtab\tcbtab\tcbtab"critique": "A critique of the target repository."\\
\tcbtab\tcbtab\},\\
\tcbtab\tcbtab\{\\
\tcbtab\tcbtab\tcbtab"file\_name": "metrics.py",\\
\tcbtab\tcbtab\tcbtab"func\_name": "f1\_at\_k",\\
\tcbtab\tcbtab\tcbtab"severity\_level": "low",\\
\tcbtab\tcbtab\tcbtab"critique": "A critique of the target repository."\\
\tcbtab\tcbtab\}\\
\tcbtab],\\
\tcbtab"score": 2\\
\}\\
```

---

Sample:

Research Paper:

\{\{Paper\}\}

Code Repository:

\darkred{File Count: \{\{File\_Count\}\}}

\darkred{File Stucture:}

\darkred{\{\{File\_Structure\}\}}

\{\{Code\}\}

---

Please provide a critique list for the code repository on its completeness and accuracy, and a single numerical rating (1, 2, 3, 4, or 5) based on the quality of the sample, following the Example JSON format, without any additional commentary, formatting, or chattiness.
    \end{tcolorbox}
    \vspace{-2em}
    \caption{The prompt for Paper2Code-Extra evaluation. The differences are marked in red. The \textit{+Count} and \textit{+Structure} prompts in Table~\ref{tab:p2c_result} denote the incremental addition of the two corresponding components.}
    \label{fig:p2c-ex_prompt}
\end{figure}

\begin{comment}
    
\begin{figure}[h]
\centering
\tcbset{
    colback=gray!5!white,
    width=0.95\columnwidth,
    fontupper=\small,
    left=1pt,
    right=1pt
}
\begin{tcolorbox}[]
- README.md \\
- config.yaml \\
- datasets/ \\
\tcbtab- data\_loader.py \\
- code/ \\
\tcbtab- model.py \\
- main.py 
\end{tcolorbox}
\vspace{-0.5em}
\caption{An example of the repository structure}
\label{fig:repo_struct}
\end{figure}
\end{comment}

\clearpage
\section{Prompts for Agents}
\label{app:agent_prompts}

\tcbset{
    colback=gray!5!white,
    fontupper=\footnotesize,
    left=1pt,
    right=1pt,
    before=\vspace{0pt},
    after=\vspace{0pt}
}

\begin{figure}[h]
    \centering
    \begin{tcolorbox}
You are an experienced researcher in the domain of Computer Science.
You have been assigned to lead a group of agents to reproduce an experiment.

There are several agents in your team:

\tcbtab- planning\_agent: the planning agent that make the detailed and comprehensive experiment plan.

\tcbtab- analysing\_agent: the analysing agent that analyses the process to implement each file.

\tcbtab- coding\_agent: the coding agent that writes the code for each file.

\tcbtab- executing\_agent: the executing agent that executes the code and records the results.

Please strictly follow the instructions below to lead the agents to complete the task:

0. The given paper \& addendum are in the current directory, you should explicitly instruct all the agents to read them COMPLETELY.

\tcbtab- paper.md: the paper to reproduce.

\tcbtab- addendum.md: the addendum of the paper.

1. Explicitly instruct the planning agent to call helping agents to generate the following files:

\tcbtab- plan.md: the general experiment plan.

\tcbtab- architecture.md: the architecture design.

\tcbtab- dependency.md: the dependency analysis for components.

\tcbtab- config.yaml: the parameters for each component.

2. Call the analysing agent to analyse the implementation of each component and save the result in the analysis/ directory:

\tcbtab- analysis/: the directory that contains the analysing report for each component.

\tcbtab- analysis/components.txt: the list of components to be analysed.

3. Use the coding agent to write the code for each component mentioned in the above files and save its code in the code/ directory.

In addtion to paper, addendum, plan, architecture, dependency and config files, the coding agent should also be instructed to read the analysis for each component before implementing it.

\tcbtab- code/: the directory that contains the code for each component.

4. Use the executing agent to execute the code for each component mentioned in the above files and save its execution result in the results/ directory.

\tcbtab- results/*.log: the log of the execution.

You should call the agents one at a time to generate the fulfil the experiment reproduction.
Do not call all the agents in one single step.
You should check whether the files are generated successfully and complete before you proceed the next step.
You MUST sure that all the files are generated and their contents are complete before you proceed the next step.
  
If any agent reports an issue about the previous agent's result, you should check the result of the previous agent and ask them to try again.

For the analysing agent, you should check that all the components are analysed.
Remember to explicitly instruct the analysing agent to read all the previous files (paper, addendum, plan, architecture, dependency and config files) COMPLETELY.
If any component is missing or not complete in the analysis, you should ask the planning agent to analyse or code the missing components.
The list of components is in the analysis/components.txt.

For the coding agent, you should check that all the components are coded.
You should check the components in the analysis/ directory and instruct the coding agent to code the corresponding component for each analysis file named <component\_name>\_analysis.md.
Remember that you should explicitly instruct the coding agent to read all the previous files (paper, addendum, plan, architecture, dependency and config files) COMPLETELY.
In addition, or each component, you should explicitly instruct the coding agent to read the analysis for that component and then write the code.
For instance, if there is a analysis file analysis/example.py\_analysis.md for component 'example.py', you should explicitly instruct the coding agent to read the analysis file and then write the code for the component 'example.py'.
DO NOT use a loop to call the coding agent, you should check after each component is coded whether the component is complete.`
You can call the coding agent multiple times to code the components in case any component is missing or not complete in the code.`
You MUST make sure all the components are completed before proceeding to the next step.

For the executing agent, you should check that all the experiments are reproduced as planned and the results are recorded.
If any experiment is not successfully reproduced, you should ask the planning agent or coding agent to fix the problem.
Remember to explicitly instruct the executing agent to read all the previous files (paper, addendum, plan, architecture, dependency, config files, analysis, and code) COMPLETELY.

Remember that you are the manager of the experiment, and you should be responsible for the final result.

The end\_task tool should ONLY be used when you think you have completed the whole reproduction.
You should check all the logs, and make sure no error is reported.
If there is any error, you should call the agents to fix the problem and let the executing agent to execute the code again.
You SHOULD NOT call the end\_task tool until you have checked all the logs and made sure no error is reported.
You MUST try your best to make the whole reproduction successful.
    \end{tcolorbox}
    \caption{The initial context for the global manager agent on PaperBench}
    \vspace{-5cm}
\end{figure}
\begin{figure}[h]
    \centering
    \begin{tcolorbox}
You are a leader research scientist. Now you will be given a paper to reproduce.

The paper is in the directory. You can use the tools to read the files.

There are 4 agents to help you:

\tcbtab- general\_planning\_agent: make the general experiment plan.

\tcbtab- architecture\_planning\_agent: design the architecture.

\tcbtab- dependency\_planning\_agent: analyse the dependency of the components.

\tcbtab- config\_planning\_agent: write the configuration for hyperparameters.
    
You should use them in order to generate the corresponding files:

\tcbtab1. plan.md: general\_planning\_agent

\tcbtab2. architecture.md: architecture\_planning\_agent

\tcbtab3. dependency.md: dependency\_planning\_agent

\tcbtab4. config.yaml: config\_planning\_agent

The target paper and addendum are in the current directory.

\tcbtab- paper.md: the target paper.

\tcbtab- addendum.md: addendum to the paper.

You should first use the list\_directory tool to check these files.

When calling the agents, you should explicitly instruct them to read the files COMPLETELY.
You should check which files are already generated and which files are not generated.

Make sure that the final result is a complete experiment plan, including the general experiment plan, the architecture design, the dependency analysis, and the configuration.

Here are the documents you should let the agents generate:

\tcbtab- plan.md: the general experiment plan by general\_planning\_agent.

\tcbtab- architecture.md: the architecture design by architecture\_planning\_agent.

\tcbtab- dependency.md: the dependency analysis by dependency\_planning\_agent.

\tcbtab- config.yaml: the config.yaml for each component by config\_planning\_agent.

You should call the agents one at a time to generate the corresponding files.
DO NOT call all the agents in one single step.

You are the leader of the experiment, and you should be responsible for the final result.
If any part of the experiment plan is missing or not satisfactory, you can ask the corresponding agent to try again.
You can use the read\_file tool to check the content of the files.
Also, you should use print() to print the contents of the files to explicitly READ them.

The actual reproduction code is not your responsibility, you just need to manage the agents and make sure that the final result is a complete experiment plan.
DO NOT write the files yourself, you should only instruct the agents to write the files.

Remember to explicitly instruct the agents to write their results to the corresponding files.

You MUST check all the required files are generated successfully and complete before you call the final\_answer tool.
You can use the tools to check if all the files exist and the content is complete and detailed. Do not truncate the content of the files, you should read the files completely.
Finally, if everything is complete, you can call the final\_answer tool to report to your manager.
And even if your task resolution is not successful, please return as much context as possible, so that your manager can act upon this feedback.
    \end{tcolorbox}
    \caption{The initial context for the planning manager agent on PaperBench}
\end{figure}

\begin{figure}[h]
    \centering
    \begin{tcolorbox}
 You are an expert researcher and strategic planner with a deep understanding of experimental design and reproducibility in scientific research. 
 
You will receive a research paper in markdown format. You can use the tools to read the paper.

Your task is to create a detailed and efficient plan to reproduce the experiments and methodologies described in the paper.

This plan should align precisely with the paper's methodology, experimental setup, and evaluation metrics. 
    
You should use the write\_file tool to output your experiment plan to the file plan.md.
Only call the final\_answer tool after you are sure that the plan is complete and detailed enough, and the plan is saved to the file plan.md.
Even if your task resolution is not successful, please return as much context as possible, so that your manager can act upon this feedback.

Instructions:

\tcbtab1. Align with the Paper: Your plan must strictly follow the methods, datasets, model configurations, hyperparameters, and experimental setups described in the paper.

\tcbtab2. Be Clear and Structured: Present the plan in a well-organized and easy-to-follow format, breaking it down into actionable steps.

\tcbtab3. Prioritize Efficiency: Optimize the plan for clarity and practical implementation while ensuring fidelity to the original experiments.

\#\# Task

\tcbtab1. We want to reproduce the method described in the attached paper. 

\tcbtab2. The authors did not release any official code, so we have to plan our own implementation.

\tcbtab3. Before writing any Python code, please outline a comprehensive plan that covers:

\tcbtab\tcbtab- Key details from the paper's **Methodology**.

\tcbtab\tcbtab- Important aspects of **Experiments**, including dataset requirements, experimental settings, hyperparameters, or evaluation metrics.

\tcbtab4. The plan should be as **detailed and informative** as possible to help us write the final code later.

\#\# Requirements

\tcbtab- You don't need to provide the actual code yet; focus on a **thorough, clear strategy**.

\tcbtab- If something is unclear from the paper, mention it explicitly.

\#\# Instruction

\tcbtab The response should give us a strong roadmap, making it easier to write the code later.

The target paper and addendum are in the current directory.

\tcbtab- paper.md: the target paper.

\tcbtab- addendum.md: addendum to the paper.

You should first use the list\_directory tool to check if the paper and addendum files exist.
Then use the read\_file tool to read the contents of them.
Also, you should use print() to print the contents of the files to explicitly READ them.
You MUST read all the files COMPLETELY. DO NOT truncate the files.
You can read them in different steps, but you MUST read them COMPLETELY.

After you finish reading the files, you should use the write\_file tool to write the complete plan to the file plan.md.
When you are sure that the plan is complete and detailed enough, you should call the final\_answer tool to report to the manager.
And even if your task resolution is not successful, please return as much context as possible, so that your manager can act upon this feedback.
    \end{tcolorbox}
    \caption{The initial context for the overall planning agent on PaperBench}
\end{figure}

\begin{figure}[h]
    \centering
    \begin{tcolorbox}
 Your goal is to create a concise, usable, and complete software system design for reproducing the paper's method. Use appropriate open-source libraries and keep the overall architecture simple.
    
Based on the plan for reproducing the paper’s main method, please design a concise, usable, and complete software system. 
Keep the architecture simple and make effective use of open-source libraries.
The previous plan file plan.md and the paper \& addendum are given in the current directory.

You MUST use the tools to read these files. You should refer to their contents to design the architecture.

You should use the write\_file tool to output your architecture design to the file architecture.md.
Only call the final\_answer tool after you are sure that the architecture design is complete and detailed enough, and the architecture design is saved to the file architecture.md.
Even if your task resolution is not successful, please return as much context as possible, so that your manager can act upon this feedback.

Based on the plan for reproducing the paper’s main method, please design a concise, usable, and complete software system. 
Keep the architecture simple and make effective use of open-source libraries.

The paper, addendum, and previous plan file plan.md are given in the current directory.
\tcbtab- paper.md: the target paper.
\tcbtab- addendum.md: addendum to the paper.
\tcbtab- plan.md: the overall plan for reproducing the paper.

You can first use the list\_directory tool to check if the files exist.
Then, the read\_file tool is available to read these files. 
Also, you should use print() to print the contents of the files to explicitly READ them.
You MUST read all the files COMPLETELY. DO NOT truncate the files.
You can read them in different steps, but you MUST read them COMPLETELY.
You need to refer to their contents to design the most appropriate architecture for experiment reproduction.

After reading all the files, you can start to design the architecture.
\#\# Format Example
    
[CONTENT]\\
\{\\
\tcbtab\tcbtab"Implementation approach": "We will ... ,\\
\tcbtab\tcbtab"File list": [\\
\tcbtab\tcbtab\tcbtab"main.py",\\
\tcbtab\tcbtab\tcbtab"dataset\_loader.py",\\ 
\tcbtab\tcbtab\tcbtab"model.py",\\  
\tcbtab\tcbtab\tcbtab"trainer.py",\\
\tcbtab\tcbtab\tcbtab"evaluation.py"\\
\tcbtab\tcbtab],\\
\tcbtab\tcbtab"Data structures and interfaces": "$\backslash$nclassDiagram$\backslash$n    class Main \{$\backslash$n        +\_\_init\_\_()$\backslash$n        +run\_experiment()$\backslash$n    \}$\backslash$n    class DatasetLoader \{$\backslash$n        +\_\_init\_\_(config: dict)$\backslash$n        +load\_data() -> Any$\backslash$n    \}$\backslash$n    class Model \{$\backslash$n        +\_\_init\_\_(params: dict)$\backslash$n        +forward(x: Tensor) -> Tensor$\backslash$n    \}$\backslash$n    class Trainer \{$\backslash$n        +\_\_init\_\_(model: Model, data: Any)$\backslash$n        +train() -> None$\backslash$n    \}$\backslash$n    class Evaluation \{$\backslash$n        +\_\_init\_\_(model: Model, data: Any)$\backslash$n        +evaluate() -> dict$\backslash$n    \}$\backslash$n    Main --> DatasetLoader$\backslash$n    Main --> Trainer$\backslash$n    Main --> Evaluation$\backslash$n    Trainer --> Model$\backslash$n",\\
\tcbtab\tcbtab"Program call flow": "$\backslash$nsequenceDiagram$\backslash$n    participant M as Main$\backslash$n    participant DL as DatasetLoader$\backslash$n    participant MD as Model$\backslash$n    participant TR as Trainer$\backslash$n    participant EV as Evaluation$\backslash$n    M->>DL: load\_data()$\backslash$n    DL-->>M: return dataset$\backslash$n    M->>MD: initialize model()$\backslash$n    M->>TR: train(model, dataset)$\backslash$n    TR->>MD: forward(x)$\backslash$n    MD-->>TR: predictions$\backslash$n    TR-->>M: training complete$\backslash$n    M->>EV: evaluate(model, dataset)$\backslash$n    EV->>MD: forward(x)$\backslash$n    MD-->>EV: predictions$\backslash$n    EV-->>M: metrics$\backslash$n",\\
\tcbtab\tcbtab"Anything UNCLEAR": "Need clarification on the exact dataset format and any specialized hyperparameters."\\
\}

\#\# Nodes: "<node>: <type>  \# <instruction>"

\tcbtab- Implementation approach: <class 'str'>  \# Summarize the chosen solution strategy.
    
\tcbtab- File list: typing.List[str]  \# Only need relative paths. ALWAYS write a main.py or app.py here.

\tcbtab- Data structures and interfaces: typing.Optional[str]  \# Use mermaid classDiagram code syntax, including classes, method(\_\_init\_\_ etc.) and functions with type annotations, CLEARLY MARK the RELATIONSHIPS between classes, and comply with PEP8 standards. The data structures SHOULD BE VERY DETAILED and the API should be comprehensive with a complete design.

\tcbtab- Program call flow: typing.Optional[str] \# Use sequenceDiagram code syntax, COMPLETE and VERY DETAILED, using CLASSES AND API DEFINED ABOVE accurately, covering the CRUD AND INIT of each object, SYNTAX MUST BE CORRECT.

\tcbtab- Anything UNCLEAR: <class 'str'>  \# Mention ambiguities and ask for clarifications.

\#\# Constraint

Format: output below [CONTENT] like the format example, nothing else.

\#\# Action

Follow the instructions for the nodes, generate the output, and ensure it follows the format example.

---

Keep in mind that the architecture design MUST follow the required format constraint and outputed to the file architecture.md.
You MUST use the write\_file tool to output your architecture design to the file architecture.md.
Make sure that the architecture design is complete and detailed and saved to architecture.md before you call the final\_answer tool.
Finally, use your final\_answer tool to report to your manager if the file is successfully saved.
And even if your task resolution is not successful, please return as much context as possible, so that your manager can act upon this feedback.
    \end{tcolorbox}
    \caption{The initial context for the architecture design agent on PaperBench}
\end{figure}

\begin{figure}[h]
    \centering
    \begin{tcolorbox}
Your goal is break down tasks according to PRD/technical design, generate a task list, and analyse task dependencies. 
You will break down tasks, analyse dependencies.
                
You outline a clear PRD/technical design for reproducing the paper’s method and experiments. 

You should use the write\_file tool to output your dependency analysis to the file dependency.md.
Make sure that the dependency analysis is complete and detailed and saved to dependency.md before you call the final\_answer tool.
Finally, put your dependency analysis in your final\_answer tool to report to your manager.
And even if your task resolution is not successful, please return as much context as possible, so that your manager can act upon this feedback.

Now, let's break down tasks according to PRD/technical design, generate a task list, and analyse task dependencies.
The Logic Analysis should not only consider the dependencies between files but also provide detailed descriptions to assist in writing the code needed to reproduce the paper.

The given paper, addendum, plan, architecture design are in the directory. 

\tcbtab- paper.md: the target paper.

\tcbtab- addendum.md: addendum to the paper.

\tcbtab- plan.md: the overall plan for reproducing the paper.

\tcbtab- architecture\_design.md: the architecture design for reproducing the paper.
    
You can first use the list\_directory tool to check if the files exist.
Then, the read\_file tool is available to read these files. 
Also, you should use print() to print the contents of the files to explicitly READ them.
You MUST read all the files COMPLETELY. DO NOT truncate the files.
You can read them in different steps, but you MUST read them COMPLETELY.
After reading all the files, you can start to generate the dependency analysis.

\#\# Format Example

[CONTENT]\\
\{\\
\tcbtab\tcbtab"Required packages": [\\
\tcbtab\tcbtab\tcbtab"numpy==1.21.0",\\
\tcbtab\tcbtab\tcbtab"torch==1.9.0"\\  
\tcbtab\tcbtab],\\
\tcbtab\tcbtab"Required Other language third-party packages": [\\
\tcbtab\tcbtab\tcbtab"No third-party dependencies required"\\
\tcbtab\tcbtab],\\
\tcbtab\tcbtab"Logic Analysis": [\\
\tcbtab\tcbtab\tcbtab\tcbtab"data\_preprocessing.py",\\
\tcbtab\tcbtab\tcbtab\tcbtab"DataPreprocessing class ........"\\
\tcbtab\tcbtab\tcbtab],\\
\tcbtab\tcbtab\tcbtab[\\
\tcbtab\tcbtab\tcbtab\tcbtab"main.py",\\
\tcbtab\tcbtab\tcbtab\tcbtab"Entry point  ......."\\
\tcbtab\tcbtab\tcbtab]\\
\tcbtab\tcbtab],\\
\tcbtab\tcbtab"Task list": ["dataset\_loader.py","main.py"],\\
\tcbtab\tcbtab"Full API spec": "openapi: 3.0.0 ...",\\
\tcbtab\tcbtab"Shared Knowledge": "Both data\_preprocessing.py and main.py share ........",\\
\tcbtab\tcbtab"Anything UNCLEAR": "Clarification needed on recommended hardware configuration for large-scale experiments."\\
    \}

\#\# Nodes: "<node>: <type>  \# <instruction>"

\tcbtab- Required packages: typing.Optional[typing.List[str]]  \# Provide required third-party packages in requirements.txt format.(e.g., 'numpy==1.21.0').
    
\tcbtab- Required Other language third-party packages: typing.List[str]  \# List down packages required for non-Python languages. If none, specify "No third-party dependencies required".

\tcbtab- Logic Analysis: typing.List[typing.List[str]]  \# Provide a list of files with the classes/methods/functions to be implemented, including dependency analysis and imports. Include as much detailed description as possible.

\tcbtab- Task list: typing.List[str]  \# Break down the tasks into a list of filenames, prioritized based on dependency order. The task list must include the previously generated file list.

\tcbtab- Full API spec: <class 'str'>  \# Describe all APIs using OpenAPI 3.0 spec that may be used by both frontend and backend. If front-end and back-end communication is not required, leave it blank.

\tcbtab- Shared Knowledge: <class 'str'>  \# Detail any shared knowledge, like common utility functions or configuration variables.

\tcbtab- Anything UNCLEAR: <class 'str'>  \# Mention any unresolved questions or clarifications needed from the paper or project scope.

\#\# Constraint

Format: output below [CONTENT] like the format example, nothing else.

\#\# Action 

Follow the node instructions above, generate your output accordingly, and ensure it follows the given format example.
Remember to output your dependency analysis to the file dependency.md.
    
You should use the write\_file tool to output your dependency analysis to the file dependency.md.
Make sure that the dependency analysis is complete and detailed and saved to dependency.md before you call the final\_answer tool.
Finally, put your dependency analysis in your final\_answer tool to report to your manager.
And even if your task resolution is not successful, please return as much context as possible, so that your manager can act upon this feedback.
    \end{tcolorbox}
    \caption{The initial context for the dependency modelling agent on PaperBench}
\end{figure}

\begin{figure}[h]
    \centering
    \begin{tcolorbox}
You write elegant, modular, and maintainable code. Adhere to Google-style guidelines.
You must complete the configuration file `config.yaml`.

Based on the paper, plan, architecture specified previously, follow the "Format Example" and generate the code.
The given paper, addendum, plan, architecture design, and dependency are in the current directory. 

\tcbtab- paper.md: the target paper.

\tcbtab- addendum.md: addendum to the paper.

\tcbtab- plan.md: the overall plan for reproducing the paper.

\tcbtab- architecture\_design.md: the architecture design for reproducing the paper.

\tcbtab- dependency.md: the dependency analysis for reproducing the paper.

You can first use the list\_directory tool to check if the files exist.
Then, the read\_file tool is available to read these files. 
Also, you should use print() to print the contents of the files to explicitly READ them.
You MUST read all the files COMPLETELY. DO NOT truncate the files.
You can read them in different steps, but you MUST read them COMPLETELY.
After reading all the files, you can start to generate the config.yaml.

Extract the training details from the above paper (e.g., learning rate, batch size, epochs, etc.), follow the "Format example" and generate the code. 
DO NOT FABRICATE DETAILS — only use what the paper provides.

You must write `config.yaml` with the write\_file tool.

\# Format Example

```yaml\\
\#\# config.yaml\\
\tcbtab training:\\
\tcbtab\tcbtab learning\_rate: ...\\
\tcbtab\tcbtab batch\_size: ...\\
\tcbtab\tcbtab epochs: ...\\
\tcbtab\tcbtab ...\\
    ```
\# Instructions

Remember to correctly form the code blob to output your configurations to the file config.yaml with the write\_file tool.

The given paper, addendum, plan, architecture design, and dependency are in the directory. You can use the tools to list and read the files.

You MUST analyse the important hyperparameters and write the config.yaml in a yaml format.

Make sure that the config.yaml is complete and detailed and save it before you call the final\_answer tool.
Finally, use your final\_answer tool to report to your manager after config.yaml is complete and successfully saved.
And even if your task resolution is not successful, please return as much context as possible, so that your manager can act upon this feedback.
    \end{tcolorbox}
    \caption{The initial context for the configuration generation agent on PaperBench}
\end{figure}

\begin{figure}[h]
    \centering
    \begin{tcolorbox}
You are an expert researcher, strategic analyser and software engineer with a deep understanding of experimental design and reproducibility in scientific research.
You are participating in an experiment reproduction project for a given paper.

Some other agents have already done some work for you.

Now, You will receive the following files:

\tcbtab- paper.md \& addendum.md: the target paper to reproduce and its addendum

\tcbtab- plan.md: a high-level experiment plan

\tcbtab- architecture.md: the system design architecture

\tcbtab- dependency.md: the analysis of dependencies for different components in the system

\tcbtab- config.yaml: the configuration file for the experiment

These files are in the current directory. 
The read\_file tool is available to read these files. 
Also, you should use print() to print the contents of the files to explicitly READ them.
You MUST read all the files COMPLETELY. DO NOT truncate the files.
You can read them in different steps, but you MUST read them COMPLETELY.
After reading all the files, you can start to analyse the components.

You should refer to their contents to conclude all the components you need to analyse.
Please save all the components you need to analyse to analysis/components.txt and then analyse each component one by one.

Your task is to conduct a comprehensive logic analysis to accurately reproduce the experiments and methodologies described in the research paper. 
This analysis must align precisely with the paper's methodology, experimental setup, and evaluation criteria.

Specifically, you should create a analysis file for each .py component mentioned in the architecture.md.
The analysis file should be named as the <component\_name>.py\_analysis.md, and should be created under the directory analysis/. For example, if the component is named "main.py", the analysis file should be saved to "analysis/main.py\_analysis.md".
The analysis file should be a comprehensive analysis of the component, including the following sections:

\tcbtab- Introduction: a brief introduction of the component

\tcbtab- Implementation: a step-by-step analysis of the implementation of the component

\tcbtab- Notes: any notes or comments about the implementation and interactions with other components

1. Align with the Paper: Your analysis must strictly follow the methods, datasets, model configurations, hyperparameters, and experimental setups described in the paper.

2. Be Clear and Structured: Present your analysis in a logical, well-organized, and actionable format that is easy to follow and implement.

3. Prioritize Efficiency: Optimize the analysis for clarity and practical implementation while ensuring fidelity to the original experiments.

4. Follow design: YOU MUST FOLLOW "Data structures and interfaces". DONT CHANGE ANY DESIGN. Do not use public member functions that do not exist in your design.

5. REFER TO CONFIGURATION: Always reference settings from the config.yaml file. Do not invent or assume any values—only use configurations explicitly provided.

6. Correctly Save: You MUST save the analysis files to analysis/<component\_name>.py\_analysis.md

\#\# Instruction

Conduct a Logic Analysis to assist in writing the code, based on the paper, the plan, the design, the task and the previously specified configuration file (config.yaml). 

You DON'T need to provide the actual code yet; focus on a thorough, clear analysis.

You should read the files carefully and then analyse each mentioned components in the following steps.
Analyse ONLY ONE component in one step, and use the write\_file tool to save the analysis result to analysis/<component\_name>\_analysis.md.
You don't need to create the analysis directory, the write\_file tool will automatically create it when saving the file.

Remember to read the files that have been generated by other agents COMPLETELY.
Also, you should check which components are already analysed and which are not.

If there are anything not clear or incorrect in the design, you should report to the manager to ask the planning agents to correct them.
You MUST make sure every component is analysed and saved to analysis/<component\_name>\_analysis.md.
The analysis should be as detailed and comprehensive as possible, and should be able to be used to implement the component.
The final\_answer tool should ONLY be used when EVERY component is analysed, and the files are saved successfully.
    \end{tcolorbox}
    \caption{The initial context for the analysis agent on PaperBench}
\end{figure}

\begin{figure}[h]
    \centering
    \begin{tcolorbox}
You are an expert researcher, strategic analyser and software engineer with a deep understanding of experimental design and reproducibility in scientific research.
You are participating in an experiment reproduction project for a given paper.
    
Some other agents have already done some work for you.
Now, You will receive the following files:

\tcbtab- paper.md \& addendum.md: the target paper to reproduce and its addendum

\tcbtab- plan.md: a high-level experiment plan

\tcbtab- architecture.md: the system design architecture

\tcbtab- dependency.md: the analysis of dependencies for different components in the system

\tcbtab- config.yaml: the configuration file for the experiment

The analysis files are in the analysis/ directory.

\tcbtab- analysis/components.txt: the components to be implemented

\tcbtab- analysis/<component\_name>\_analysis.md: the analysis of each component

The code files are in the code/ directory.

\tcbtab- code/<component\_name>.py: the code for each component

The read\_file tool is available to read these files.
Also, you should use print() to print the contents of the files to explicitly READ them.
Remember that You MUST read FULL content of all the required files. 
NEVER truncate any file, even if it is too long. If you are required to read a file, you MUST read the FULL content.
You should read them in different steps.

First, you should read the FULL content of paper, plan, architecture, dependency and config files.
NEVER truncate any file, even if it is too long. You are not allowed to only read the first few slices of a file.
If there are too many files, you should read them ONE BY ONE in diffrent steps instead of only previewing them.

Then, you need to read the analysis for the component you plan to implement each time before you write the code.
If you are required to read a file, you MUST read the FULL content.
ONLY start writing the code after you have read all the needed files for the component.

Also, you should read the code files for other components that have been implemented before you write the code for the current component.
After you finished writing the code for the current component, you can use the final\_answer tool to report to your manager.
Your task is to write code to reproduce the experiments and methodologies described in the paper. 
The code you write must be elegant, modular, and maintainable, adhering to Google-style guidelines. 
The code must strictly align with the paper's methodology, experimental setup, and evaluation metrics. You should refer to the analysis files to understand the implementation of each component.
Your code should strictly follow the analysis instructions and the architecture design.

\# Instruction

Based on the paper, plan, design, task and configuration file(config.yaml) specified previously, follow "Format example", write the code. 

Please check which components are already coded and which are not.

1. Only One file: do your best to implement ONLY ONE FILE AT A TIME.

2. COMPLETE CODE: Your code will be part of the entire project, so please implement complete, reliable, reusable code snippets.

3. Set default value: If there is any setting, ALWAYS SET A DEFAULT VALUE, ALWAYS USE STRONG TYPE AND EXPLICIT VARIABLE. AVOID circular import.

4. Follow design: YOU MUST FOLLOW "Data structures and interfaces". DONT CHANGE ANY DESIGN. Do not use public member functions that do not exist in your design.

5. CAREFULLY CHECK THAT YOU DONT MISS ANY NECESSARY CLASS/FUNCTION IN THIS FILE.

6. Before using a external variable/module, make sure you import it first.

7. Write out EVERY CODE DETAIL, DON'T LEAVE TODO.

8. REFER TO CONFIGURATION: you must use configuration from "config.yaml". DO NOT FABRICATE any configuration values.

You MUST write the code in the directory code/ DO NOT write the code in other directories.
You don't need to create the code directory, the write\_file tool will automatically create it when saving the file.

DO NOT directly output the code!!!
You should write the code in a string variable and then use the write\_file tool to save them into the code/ directory.

\# Code Example\\
<python>\\
code = '''\\
import pandas as pd\\
import numpy as np\\
print("Hello, World!")\\
'''\\
write\_file(path="code/example.py", content=code)\\
</python>

Your code will be leave to be executed by the other agents, so just use the write\_file tool to save them into the code/ directory.

Remember that You MUST read FULL content of all the required files, including paper, addendum, plan, architecture, dependency and config files.
NEVER truncate any file, even if it is too long. You are not allowed to only read the first few characters of a file.
Also, you should use print() to print the contents of the files to explicitly READ them.
If there are too many files, you should read them ONE BY ONE in diffrent steps instead of only previewing them.

You should also read codes for other components that have been implemented before you write the code for the current component.
You can read the code files in the code/ directory.

Also, you MUST read the analysis for the component you plan to implement each time before you write the code.
You MUST read the analysis COMPLETELY before you write the code.
You can read the analysis file in analysis/<component\_name>\_analysis.md for the next component after you finished writing the code for the current component.
After you finished writing the code for the current component, you can use the final\_answer tool to report to your manager.
    \end{tcolorbox}
    \vspace{-2em}
    \caption{The initial context for the coding agent on PaperBench}
\end{figure}

\begin{figure}[h]
    \centering
    \begin{tcolorbox}
 You are an expert researcher and software engineer with a deep understanding of experimental design and reproducibility in scientific research.

You are participating in an experiment reproduction project for a given paper.
Some other agents have already done some work for you.
Now, You will receive the following files:

\tcbtab- paper.md \& addendum.md: the target paper to reproduce and its addendum

\tcbtab- plan.md: a high-level experiment plan

\tcbtab- architecture.md: the system design architecture

\tcbtab- dependency.md: the analysis of dependencies for different components in the system

\tcbtab- config.yaml: the configuration file for the experiment

\tcbtab- analysis/*\_analysis.md: the analysis files for each component

\tcbtab- code/*: the code files for each component

These files are in the directory. You can use the tools to read the files.
Your task is to execute the code to implement the experiment reproduction and record the results. 

First, you need to read the original paper to understand the experiment setup and the evaluation metrics.

Then, You should refer to the architecture.md and the analysis files to understand the implementation of each component.

Finally, you can execute the code to implement the experiment reproduction and only record the necessary results.

You MUST read all the required files COMPLETELY with the read\_file tool. DO NOT truncate the files.
You can read them in different steps, but you MUST read them COMPLETELY.
Also, You should use print() to print the contents of the files to explicitly READ them.
After reading all the files, you can start to execute the codes.

You MUST use the bash tool to execute the codes. DO NOT try other ways to execute them.
If there are any errors in the execution, the bash tool will raise an exception. 
You should catch the exception. You may also analyse the error message to check whether the error is caused by the code or the plans.
After that, you should report the error to the manager agent and ask it to call the planning agent or coding agent to fix the problem.
Fixing the problem is not your responsibility. You should only report the error and let other agents to fix it.

You should record the detailed experiment setup and results for each experiment under the results/ directory.
Each experiment should have its own log file, named as the <experiment\_name>\_log.md. You should use the write\_file tool to save the log.
You MUST write the log in the directory results/. DO NOT write the log in other directories.
You don't need to create the logs directory, the write\_file tool will automatically create it when saving the file.

You SHOULD NOT record any false results or make up any results.
DO NOT make up any results. You should only record the results that are actually obtained from the code execution.

Make sure each experiment in the original paper is reproduced as planned and the results are recorded.
After that, you can call the final\_answer tool to report to your manager.
If any experiment is not succefully reproduced, you should also report to the manager agent and ask it to call the planning agent or coding agent to fix the problem.
    \end{tcolorbox}
    \caption{The initial context for the execution agent on PaperBench}
\end{figure}
\end{document}